\documentclass[preprint,11pt]{elsarticle}
\usepackage{fullpage}

\makeatletter
\def\ps@pprintTitle{%
 \let\@oddhead\@empty
 \let\@evenhead\@empty
 \def\@oddfoot{\centerline{\thepage}}%
 \let\@evenfoot\@oddfoot}
\makeatother

\usepackage[utf8]{inputenc}
\usepackage[english]{babel}

\usepackage{csquotes}

\usepackage{amsmath,amssymb,amsfonts,amsthm}
\usepackage{algorithmic}
\usepackage{graphicx}
\usepackage{textcomp}
\usepackage{wrapfig}
\usepackage{hyperref}

\usepackage[ruled,vlined]{algorithm2e}

\usepackage{tabularx}

\usepackage{enumerate}
\usepackage{indentfirst}
\usepackage{latexsym}
\usepackage{multicol}
\usepackage{verbatim}
\usepackage{color}
\usepackage{parallel}
\usepackage{cases}
\usepackage{array}
\usepackage{booktabs}
\usepackage{framed}

\usepackage{booktabs}
\usepackage[usenames,dvipsnames]{xcolor}
\usepackage{tikz-cd}
\usetikzlibrary{calc,arrows,positioning}
\usepackage{pgfplots}
\pgfplotsset{compat=1.14}
\usepackage{graphicx}

\newcommand{\R}{{\mathbb R}}

\def\cN{{\mathcal N}}
\def\A{{\mathcal A}}
\newcommand{\vx}{\mathbf x}

\newtheorem{defi}{Definition}
\newtheorem{lem}{Lemma}

\newtheorem{prop}{Proposition}
\newtheorem{theo}{Theorem}

\usepackage{MnSymbol}
\newtheorem{remark}{Remark}

\begin{document}

\begin{frontmatter}
\title{Data Clustering as an Emergent Consensus of Autonomous Agents}
\author[add1]{Piotr Minakowski}
\ead{piotr.minakowski@ovgu.de}
\author[add2]{Jan Peszek}
\ead{j.peszek@mimuw.edu.pl}


\address[add1]{\footnotesize Institute of Analysis and Numerics, Otto von Guericke University Magdeburg, Universit\"atsplatz 2, 39106 Magdeburg, Germany}
\address[add2]{\footnotesize Institute of Applied Mathematics and Mechanics, University of Warsaw, ul. Banacha 2, 02-097 Warszawa, Poland}

\begin{abstract}
We present a data segmentation method based on a first-order 
density-induced consensus protocol.
We provide a mathematically rigorous analysis of the consensus model 
leading to the stopping criteria of the data segmentation algorithm. To 
illustrate our method, the algorithm is applied to two-dimensional 
shape datasets and selected images from Berkeley Segmentation Dataset. The 
method can be seen as an augmentation of classical clustering techniques for 
multimodal feature space, such as DBSCAN. It showcases a curious connection 
between data clustering and collective behavior.
\end{abstract}

\date{}
\end{frontmatter}

\section{Introduction}\label{sec:intro}
The paper explores the relationship between first-order collective dynamics and 
data clustering, which we illustrate on color image segmentation. 
Our main contributions 
include:
\begin{itemize}
    \item An augmentation of the DBSCAN clustering method that reduces numerical complexity.
    \item The technique inherits the advantages of other density-based clustering methods, with lower average numerical complexity and manageable parameters. 
    \item Mathematically rigorous analysis of the model leading to the stopping 
          criteria for the algorithm.
    \item Application of the method in color image segmentation.
\end{itemize}

Our approach is based on the recently introduced density-induced 
consensus protocol (DI protocol) for agent-based collective dynamics 
\cite{MiMuPe2020}. Consider $N$ agents with $x_i(t)\!\in\!\R^d$ 
denoting the position of $i$th agent in a $d$-dimensional space at the time 
$t\geq 0$. The agents follow the DI protocol
\begin{equation}{\label{1st}}
  \dot x_i= \displaystyle \kappa\sum_{k\in\cN_i}(x_k-x_i),\quad x_i(0) = 
  x_{i0}\in\R^d.
\end{equation}
\noindent
Here $\kappa>0$ is a fixed coupling strength whose influence amounts, in practice, to time-scaling. The neighbor set $\cN_i$ of $i$th agent is defined through the following 
relation: given positive parameters $\delta$ and $m$, for $t\geq 0$ we 
define,
\begin{equation}\label{nor}
\begin{split}
k\in &\ \cN_i(t)\ \Leftrightarrow\   x_k(t)\in B(x_i(t),\delta)\ \mbox{and}\ 
\\ & 
\#\Big\{k\in\{1,...,N\}: x_k(t)\in B(x_i(t),\delta)\Big\} >  m,
\end{split}
\end{equation}
where $B(x_i(t),\delta)$ is an open ball centered at $x_i(t)$ with 
radius~$\delta$ and $\# A$ denotes the cardinal number of $A$. Thus the 
communication rule of the DI protocol reads as follows: 
the~$i$th agent is influenced by $j$th agent if
\begin{itemize}
    \item the density of agents in close proximity to the $i$th agent is substantial 
    enough (otherwise the $i$th agent is an outlier),
    \item the $j$th agent is in close proximity to the $i$th agent.
\end{itemize}

\smallskip

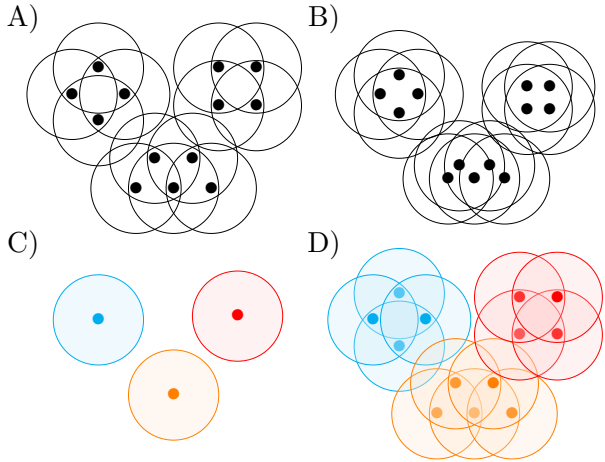
\begin{figure}[h!]
\begin{tikzpicture}[x=.5cm,y=.5cm]
  \node at (0,12) {A)};
  \node at (8,12) {B)};
  \node at (0,6) {C)};
  \node at (8,6) {D)};
 \foreach \Point in { (2,10.7), (2,9.3), (1.3,10), (2.7,10)}{
 \node at \Point {\textbullet};
 \draw[line width=.1pt]  \Point circle (1.2);
 }

 \foreach \Point in { (3,7.5), (4,7.5), (5,7.5), (3.5,8.3),(4.5,8.3)}{
  \node at \Point {\textbullet};
 \draw[line width=.2pt]  \Point circle (1.2);
  }

  \foreach \Point in { (5.2,9.7), (6.2,9.7), (5.2,10.7), (6.2,10.7)}{
  \node at \Point {\textbullet};
  \draw[line width=.2pt]  \Point circle (1.2);
  }
 \foreach \Point in { (10,10.5), (10,9.5), (9.5,10), (10.5,10)}{
  \node at \Point {\textbullet};
  \draw[line width=.2pt]  \Point circle (1.2);
  }
\foreach \Point in { (11.3,7.75), (12,7.75),(12.8,7.75),(11.6,8.1),(12.4,8.1)}{
  \node at \Point {\textbullet};
  \draw[line width=.2pt]  \Point circle (1.2);
 }

\foreach \Point in { (13.4,9.6), (14.,9.6), (13.4,10.2), (14.,10.2)}{
  \node at \Point {\textbullet};
  \draw[line width=.2pt]  \Point circle (1.2);
 }
\foreach \Point in { (2,4)}{
  \draw[cyan,fill=cyan!20!white,fill opacity=0.25, line width=.2pt]  \Point 
  circle (1.2);
  \node[cyan] at \Point {\textbullet};
}
\foreach \Point in { (4,2)}{
  \draw[orange,fill=orange!20!white,fill opacity=0.25, line width=.2pt]  \Point 
  circle (1.2);
  \node[orange] at \Point {\textbullet};
}

\foreach \Point in { (5.7,4.1)}{
  \draw[red,fill=red!20!white,fill opacity=0.25, line width=.2pt]  \Point 
  circle (1.2);
  \node[red] at \Point {\textbullet};
}
\foreach \Point in { (10,4.7), (10,3.3), (9.3,4), (10.7,4)}{
  \draw[cyan,fill=cyan!20!white,fill opacity=0.25, line width=.2pt]  \Point 
  circle (1.2);
  \node[cyan] at \Point {\textbullet};
}
\foreach \Point in { (11,1.5), (12,1.5), (13,1.5), (11.5,2.3),(12.5,2.3)}{
  \draw[orange,fill=orange!20!white,fill opacity=0.25, line width=.2pt]  \Point 
  circle (1.2);
  \node[orange] at \Point {\textbullet};
}

\foreach \Point in { (13.2,3.6), (14.2,3.6), (13.2,4.6), (14.2,4.6)}{
  \draw[red,fill=red!20!white,fill opacity=0.25, line width=.2pt]  \Point 
  circle (1.2);
  \node[red] at \Point {\textbullet};
}
\end{tikzpicture}
 \caption{  An ensemble of $13$ agents with visualisation of their range of interactions. A) initial positions of the agents; B) positions evolve towards the 
  steady-state; C) steady state at $t=\infty$ defines the clusters (color 
  coded); D) 
  clustering applied to the initial state. }
\label{fig1}
\end{figure}

Interpreting the positions $x_i$ as data points in a multimodal feature space 
and evolving the data in time using \eqref{1st} results in a 
data clustering. 
To explain this idea, here we 
fix our attention to 2D data represented as vectors/positions in a Euclidean 
space. The positions $x_i(t)\!\in\!\R^2$ for $i\in\{1,...,N\}$ evolve according 
to~\eqref{1st} eventually 
leading to a steady state at the time $t=\infty$, typically forming multiple 
clusters. Finally, the $t=\infty$ clustering is retroactively applied to the 
initial $t=0$ state, to establish the segmentation, see Fig.~\ref{fig1}.

It is noteworthy that the communication rule of \eqref{1st} is equivalent to 
the density-based spatial clustering algorithm (DBSCAN) \cite{Ester1996} widely 
used in data segmentation \cite{Schubert2017}. Our method inherits the 
advantages (and most disadvantages) of the classical density-based clustering 
algorithms  {\it i.e.}:  
no requirement of initial specification of the number of clusters, arbitrary 
shaped clusters, robustness to outliers and adaptability to various types of 
data due to the possibility of fine-tuning the parameters. 
The main difference is that we do not need database-oriented range-queries, 
which leads to the computational complexity of DBSCAN. Instead, our procedure 
is based on local direct connectivity defined by the neighbor sets ${\mathcal 
N}_i$, and thus it is less computationally demanding. 

Concerning the above, the DI protocol applies to consensus dynamics as 
discussed in \cite{MiMuPe2020}. In particular, it is reminiscent of the 
bounded-confidence consensus model by Hegselmann and Krause 
\cite{Hegselmann2002}. Similarly, the agents are influenced only by like-minded 
({\it i.e.} nearby) 
individuals but in the case of the DI model, the agents additionally tend to 
conform to group mentality and ignore the outliers.
Since our method originates in consensus dynamics it can be particularly useful 
in segmentation of 
data related to collective dynamics such as the classification of pedestrians 
or the users of social
networks. It is in itself an interesting application of non-local collective 
dynamics in data clustering.

The paper is organized as follows. In Section \ref{sec:prev} we compare the DI 
protocol to previous research on consensus dynamics and density based 
clustering, focusing on unsupervised color image segmentation.
In~Section \ref{sec:dynamics} we provide an analysis of 
emergent behavior of the DI protocol based on the interplay between density of 
the agents and connectivity. Sections \ref{sec:alg} and \ref{sec:2dexamples} 
are dedicated to the presentation of the main segmentation algorithm and 
its illustration on classical data sets, respectively. Here, we also apply the 
results of Section \ref{sec:dynamics} to derive the~stopping criterion. Section \ref{sec:appl} focuses on applications in color image segmentation.  Finally, 
in Sections \ref{sec:math} and \ref{sec:conc} we provide mathematical proofs 
related to the results of Section \ref{sec:dynamics}, and a brief conclusion 
to the paper, respectively.

{\bf Notation.} Throughout the paper we will sometimes interchangeably 
refer to agents following the DI protocol as nodes, data or pixels 
depending on the context of applications.

\section{Previous research}\label{sec:prev}

\subsection{Consensus dynamics}

Protocol (\ref{1st}-\ref{nor}) is a first-order variant of the DI model 
introduced in \cite{MiMuPe2020} which simulates density based alignment in the 
spirit of the Cucker-Smale model (\cite{CS}, see also surveys \cite{CSsurv1} and \cite{CSsurv2}). 
First-order models similar to \eqref{1st} with a variety of interaction laws 
have been extensively studied from the perspective of both mathematical theory 
and practical applications, with the prominence of applications in 
opinion dynamics. Opinion dynamics dates back to French's research on social 
influence \cite{FREAFT}, and further, to works by De Groot \cite{GROOT} and Lehrer 
\cite{LEHRER}. A more modern approach, including nonlinear models, was 
established by 
Krause \cite{KRAUSE} and Hegselmann jointly with Flache \cite{F-H}, among 
others. We emphasise the 
aforementioned work by Hegselmann and Krause \cite{Hegselmann2002} introducing 
a bounded confidence opinion dynamics model. We also recommend surveys \cite{OPsurv1} and \cite{OPsurv2} for 
more up-to-date information, and \cite{Surv3} where similar models are presented in applications to vehicular traffic and crowd dynamics.

\subsection{Density based clustering}\label{sec:dbclust}

Since most of the clustering techniques \cite{Aggarwal2013}
are application dependent, they are not tailored to the
classification of arbitrary feature space.
Methods that rely explicitly on the number of clusters or 
implicitly assume the same shape for
all the clusters are not well suited to analyse 
data of unknown origin, see \cite{Jain2000} for a survey.

Density based clustering is a class of unsupervised learning methods. 
Points that are in high-density regions in a 
data space are clustered together and separated from other such clusters 
by regions of low point density \cite{Herbin1996,Kriegel2011}.

As already mentioned in the introduction, we group points in a similar way 
as DBSCAN \cite{Ester1996} does.
The OPTICS \cite{Ankerst1999} algorithm is an extension of DBSCAN and 
removes its dependence on parameters. Another related iterative technique 
is Mean-shift algorithm \cite{Cheng1995}, where each object is assigned 
to the densest area in its vicinity, based on kernel density estimation.

Unsupervised image segmentation is an important component in many 
image processing systems. A nonparametric technique for the analysis 
of a complex multimodal feature space based on mean-shift algorithm 
with application to image segmentation have been presented in 
\cite{Comaniciu2002}.
On the other hand, one can apply AI based approaches. 
The usage of convolutional neural networks for unsupervised image segmentation
have been studied in \cite{Kanezaki2018} or \cite{Kim2020}.
Note that, image segmentation is an ill-defined problem since there is no 
unique ground-truth segmentation of an image against which the output of an 
algorithm may be compared. For the evaluation methods, we refer to 
\cite{Unnikrishnan2007} and survey \cite{Zhang2008}.

The connection between clustering and collective dynamics,
has been already explored in the context of particle swarm optimization 
(PSO)~\cite{Eberhart1995}. For density based PSO clustering algorithms we refer 
to \cite{Alswaitti2018} and \cite{Guan2019}.
Moreover, PSO based image clustering has been developed in~\cite{Omran2005}. 
The only similarity between PSO and our method is the agent based approach 
itself. 
However the rules of the agents' evolution are different, and lead to different 
characteristics of the methods, {\it e.g.} PSO requires to {\it  a priori} 
specify the number of clusters, 
and is based on solving global optimisation problem -- a computationally 
expensive process. On the other hand, we utilise a simple rule for the 
evolution of agents \eqref{1st} allowing us to employ techniques of linear 
consensus theory \cite{OLF1, OLF2}.

\section{Emergent dynamics of the DI protocol}\label{sec:dynamics}
In this section we analyse the long time dynamics of the DI protocol, focusing 
on cluster formation.  Before we begin, let us introduce the necessary notation.

The DI consensus protocol \eqref{1st} can be naturally represented in the 
language of graph theory. 
The ensemble $\{1,...,N\}$ is identified with a directed graph (digraph) 
$(\{1,...,N\},{\mathcal E})$ with edges ${\mathcal E}$ defined by neighborhoods 
$\cN_i$: 
\[{\mathcal E}(i,j)=1\ \Leftrightarrow\ i\in \cN_j,\]
cf. \eqref{nor}. Note that, since $i\in\cN_j$ 
does not imply $j\in\cN_i$, the matrix ${\mathcal E}$ is not necessarily 
symmetric, and thus the graph $(\{1,...,N\},{\mathcal E})$ is directed. 
We shall refer to {\it all} subsets of $\{1,...,N\}$ as {\it clusters}.
What follows, is a natural inheritance of standard notions from graph theory; 
for instance a cluster $\A\subset \{1,...,N\}$  is { \it weakly/strongly 
connected} iff its respective graph is weakly/strongly connected. We will also 
say that a cluster $\A\subset\{1,...,N\}$ is {\it isolated} iff it is not 
connected to any node outside of $\A$.

Next we provide the essential
notion of a densely-packed cluster introduced in \cite{MiMuPe2020}.
\smallskip
\begin{defi}\label{packed}
    {\it We say that the cluster $\A\subset\{1,...,N\}$ is \mbox{$r$-densely} 
    packed  at the time $t$ if
    \begin{enumerate}
        \item the set $$\bigcup_{i\in\A}B(x_i(t),r/2)$$ is connected,
        \item each open ball $B(x_i(t),r)$, for $i\in \A$ contains more than $m$ agents.
    \end{enumerate}} 
\end{defi}
\smallskip

There are multiple connectivity- and graph-related implications of Definition \ref{packed}. 
Most of them, are presented in Section 
\ref{sec:math}, Lemma \ref{con}. The main application is the following theorem (also proved in 
Section \ref{sec:math}) on sufficient conditions ensuring that the 
clusters collapse into a single steady state.

\smallskip
\begin{theo}\label{theo1}
{\it
Suppose that at any time $t_0$ the ensemble $\{1,...,N\}$ consists of $K$ isolated connected
clusters $\A_1,...,\A_K$ with the property that the convex hulls of $\delta/2$-neighborhoods of the clusters do not intersect, {\rm i.e.}
\begin{equation}\label{wypuk}
 {\rm conv}\left(\bigcup_{i\in\A_k}B(x_i,\delta/2)\right)
    \cap {\rm conv}\left(\bigcup_{i\in\A_l}B(x_i,\delta/2)\right) = \emptyset
\end{equation}
for all $k,l\in\{1,...,K\}$. Then \eqref{wypuk} persists in 
time and the clusters remain isolated. Moreover, 
each $r$-densely packed 
cluster $\A$ with $r$
satisfying 
\begin{align}\label{warunek}
  \frac{r}{\delta}{\mathcal Z}^\frac{r}{\delta}\leq \frac{m}{6\#\A} {\mathcal Z},\quad \quad {\mathcal Z} := e^\frac{2m}{3(\#\A)^3}
\end{align}
is $\delta$-densely packed indefinitely, and converges to the steady state 
$$x_\A:=\frac{1}{\#\A}\sum_{i\in\A}x_i.$$
}
\end{theo}

\smallskip

\begin{remark}
  Condition \eqref{warunek} may not be clear at the first glance, but is easily 
  viable from the perspective of numerical computations. Intuition behind it is 
  as follows. If $\A$ is $r$-densely packed with $r\leq\delta$, then it is strongly connected (see Lemma \ref{con} in Section \ref{sec:math} below). Consequently, if $r$ is significantly smaller than $\delta$, the agents have room to move around, before the clusters ceases to be at least $\delta$-densely packed. Thus, the quantity $\frac{r}{\delta}$ represents the cluster's rigidity, {\it i.e.} small $\frac{r}{\delta}$ means that the cluster is 
  flexible and the agents may move a lot before connectivity breaks. 
  Condition \eqref{warunek} requires rigidity to be small compared to 
  quantities related to the cluster's volume and the algebraic connectivity 
  for \eqref{1st}.

In practice we usually have $2m\ll 3(\#\A)^3$, for which
$${\mathcal Z} \approx 1$$
and then \eqref{warunek} can be reduced to
\begin{equation}\label{rrr}
    r \approx \frac{\delta m}{6\#\A}.
\end{equation}
We use simplification \eqref{rrr} in numerical computation.
\end{remark}

\begin{remark}
  Theorem \ref{theo1} states that as soon as convex hulls of the clusters' 
  influence regions are disjoint and any cluster is sufficiently densely 
  packed, then, from the  
  perspective of large-time behaviour, that cluster can be immediately replaced 
  by its center of mass $x_\A$. It naturally leads to a stopping criterion 
  for the clustering algorithm presented in Section \ref{sec:stop} below.
\end{remark}

\section{Algorithm}\label{sec:alg}

The continuous model presented in Section~\ref{sec:intro} can be simulated with 
a time-discretization scheme. This results in an iterative procedure that, for a 
given set of agents, performs clustering by advancing their positions in time up 
to the point when a sufficiently densely packed configuration is reached. The procedure 
is presented in Algorithm~\ref{alg} with a detailed description of the key steps 
in sections \ref{sec:stop} - \ref{sec:prepost}.

\begin{algorithm}\label{alg}
  \SetAlgoLined
  \KwIn{ data(0) and parameters ($\delta$, $m$, $n_{max}$)}
  Pre-processing (optional)\;
  Build $\delta-$lattice\;
  \For{$n=0$; $n<n_{max}$; $n++$}{
    Interactions(data(n), $\delta$, m)\;
    Advance the data in time\;
    }
  Build $\epsilon-$lattice\;
  Identify clusters\;
  Assign colors and outliers to clusters (optional)\;
  Post-processing (optional)\;
  \caption{Segmentation Procedure.}
\end{algorithm}

\subsection{Input}\label{sec:stop}

The input consists of the given data(0) scaled to the unit cube $[0,1]^d$,  
parameters $\delta$, $m$,
and the stopping time $n_{max}$. Parameters $\delta$ and $m$ are chosen 
to fit particular types of data, similar to other unsupervised data 
segmentation methods. The influence of parameters is investigated 
in Section~\ref{sec:params}.

In order to fix the stopping time $n_{max}$ we test Algorithm~\ref{alg} on 
sample sets of data. We perform the loop from Algorithm~\ref{alg} until all 
clusters satisfy \eqref{wypuk} and are $r$-densely packed with $r$ 
satisfying~\eqref{rrr}. 
Then we set $n_{max} = $ {\it the number of the last iteration}. 
This process is computationally expensive, but it needs to be performed 
only once for each type of similar data. For example, in all applications 
presented in the sequel, we take $n_{max} = 10$.

\subsection{Build $\delta-$lattice}
By \eqref{1st} and \eqref{nor} the communication protocol is 
local and depends on interaction range $\delta$. 
Therefore, we introduce the division of $[0,1]^d$  
into a~regular lattice of $N_l^d$ cells, with the edge length $N_l =1/\delta$.
This serves the purpose of reducing the complexity of neighbor queries, in 
order to compute the right hand side of \eqref{1st}.
The agents within $\delta-$distance are to be found in adjacent cells.
Thus, for uniformly distributed data, the complexity scales linearly 
with the number of agents. This method is applied, for instance, in particle 
simulations of liquids, c.f.~\cite{Allen2017}.

\subsection{Interactions}\label{sec:inter}
For a fixed $n\in\{0,...,n_{max}\}$, we establish the connectivity between the 
agents at the $n$th iteration, ${\it i.e.}$ the neighbor sets $\cN_i(n)$. It 
suffices to check conditions \eqref{nor} for each $i\in \{1,...,N\}$ and each 
$k$ belonging to the same $\delta$-lattice cell as $i$ or to the adjacent cells.

\subsection{Advance the data in time}
For temporal discretization we use a scheme from the Runge–Kutta 
family of explicit methods. We advance in 
time with a well-known explicit Euler method
$$ x_i^{n+1} = x_i^n + (\Delta t)\kappa \sum_{k\in\cN_i(n)}(x^n_k-x^n_i),\quad 
x_i^0 = x_{i0}.$$
Here $\cN_i(n)$ is the neighbor set obtained in the previous step. We take 
$(\Delta t)\kappa = 1/K_n$, where $$K_n = 
\max_i\Big\{\max\{\#\cN_i(n),m\}\Big\}.$$

Naturally, other choices of $(\Delta t)\kappa$ are possible, but one should 
make sure that the time-scale is small, whenever the maximal local density of 
the agents is large.

\subsection{Identify clusters}\label{sec:dpl}

After completing the $n_{max}$ iteration we construct an~$\epsilon$-lattice, for some $\epsilon\leq\delta$. 
 Note that, the maximal iteration $n_{max}$ has been chosen 
so that we expect all isolated connected clusters to be $r$-densely packed with $r$ satisfying 
\eqref{rrr}. In such a case the $\epsilon$-lattice cells emulate $r$-densely packed clusters. The cluster assignment 
is performed as follows:
\begin{enumerate}
    \item Cells of the $\epsilon$-lattice are divided into 2 groups: {\it core cells} with more than $m$ agents inside and {\it outlier cells} with $m$ or less agents inside.
    \item Any core cell forms a cluster along with all of adjacent core cells and, transitively, all subsequent adjacent core cells {\it etc.}.
    \item Agents belonging to core cells are then assigned to the respective clusters.
    \item Agents belonging to the outlier cells become outliers and are considered noise.
\end{enumerate}
Finally the cluster assignment is retroactively applied to the initial values {\it i.e.} to data(0), as showcased in Fig. \ref{fig1}.

\begin{remark}
  The scale $\epsilon>0$ of the $\epsilon$-lattice chosen to be much greater than the density threshold \eqref{rrr}, i.e. $r\ll \epsilon$. However, at $n_{max}$ we expect all of the connected 
  isolated clusters to be $r$-densely packed and asymptotically stable, by 
  Theorem \ref{theo1}. Therefore the only 
  inadequacy introduced by the procedure described in Section \ref{sec:dpl}, 
  compared to what we 
  would obtain with a finer $r$-lattice, amounts to the merging of clusters. 
  The reasonable choice of $\epsilon$ is in the same order of magnitude as 
  $\delta$. In the forthcomming examples we take either 
  $\epsilon=\frac{\delta}{2}$ or $\epsilon=\frac{\delta}{4}$, since these 
  values  provide a good balance between numerical 
  complexity, accuracy of the algorithm and resistance to the curse of 
  dimensionality. 
\end{remark}

\subsection{Assign colors and outliers to clusters}\label{sec:ass}
Optionally, for instance in the applications to color image segmentation, at 
this point we assign values (colors) to each of the clusters obtained in Step 
\ref{sec:dpl}. To do so, we simply take the average value of the data within 
each cluster. Thus, if ${\mathcal A}$ is one of the clusters, it is assigned 
with the average value 
$x_{\mathcal A} = \sum_{i\in{\mathcal A}}x_i/\#{\mathcal A}$. Then, each 
outlier cell (and the agents within) is assigned to the 
cluster with the closest average value $x_{\mathcal A}$.

In the case of color image segmentation, this step is responsible for defining the color scheme and dealing with the outliers.

\subsection{Complexity}\label{sec:complex}

For each time-step the most computationally expensive part of the algorithm is 
to establish the connectivity between the agents described in Section \ref{sec:inter}. 
Taking advantage of the $\delta$-lattice structure, for each agent we need to 
examine neighbors in adjacent lattice cells. In that case, only a limited 
number of agents must be visited multiple times, therefore the average 
complexity of 
$O(N)$ is obtained. Moreover, we perform constant number of iterations, since 
$n_{max} \ll N=$ \textit{number of agents}.

As described in Section \ref{sec:dpl}, identification of clusters is based on the $\epsilon$-lattice. We denote by $k$ the number of core cells in the $\epsilon$-lattice. Consequently, identifying clusters procedure requires $O(k \log k)$ operations. 
Therefore an overall average runtime complexity of $O(N)+O(k \log k)$ is 
obtained. If
\begin{equation}\label{kkk}
   k \log k \leq N, 
\end{equation}
then we obtain the complexity of order $O(N)$. Thanks to the fact that we only 
need to work with core cells, we can estimate $k$ in terms of $\delta$, 
regardless of the choice of $\epsilon$.
By the definition of the core cells, $k m\leq N$ and assuming uniform distribution of the agents, i.e. $\frac{m}{\delta^d}\approx \frac{N}{1}$ leads to $k\lesssim \frac{1}{\delta^d}$. Thus \eqref{kkk} is ensured if
\begin{equation}\label{ddd}
    \frac{1}{\delta^d} \log \frac{1}{\delta^d}\leq N.
\end{equation}

This is the key point of our contribution, namely the development of a method of 
low complexity, c.f.~\cite{Xu2015}. It outperforms mid complexity methods, 
{\it e.g.} DBSCAN, whose complexity is driven by distance queries. Even if an 
indexing structure is used, a neighbourhood query  executes in $O(\log N)$ 
leading to an overall average complexity of DBSCAN of order $O(N\log N)$,  with the worst 
case of $O(N^2)$,  c.f. \cite{Gan2015}.

\begin{remark}\label{imrpo}

 The improvement of the algorithm's numerical complexity hinges on assumption 
 \eqref{kkk}, which we achieve by ensuring that $\delta$ satisfies \eqref{ddd} 
 for any fixed $N$. It creates a lower bound $\delta_0(N,d)$ for the set of 
 admissible parameters $\delta$.

 Observe, that taking $n_{max} = 0$, {\it i.e.} immediately skipping to step 
 \ref{sec:dpl} of the algorithm, we actually perform a simplified variant of 
 the DBSCAN segmentation with $\epsilon$ as the neighborhood size. However, in 
 a  high-dimensional case $\delta_0(N,d)$ can be relatively large resulting in 
 far  too few clusters. In other words, proper clustering with the DBSCAN 
 algorithm requires $\delta$ to be small enough, so that we cannot ensure 
 \eqref{kkk} and we do not gain much in terms of numerical complexity. We 
 visualize this phenomenon in the case of color image 
 segmentation in Fig. \ref{fig2}. Indeed, the image in column C and row 1 
 presents such a scenario: Algorithm \ref{alg} performed with $n_{max} = 0$. It 
 leads to the emergence of only 1 cluster and a poor representation of the 
 original image in column A and row 1.
 It indicates that the key point of our method, {\it i.e.}
 the evolution of the agents' positions in time, which breaks large clusters, 
 is essential.

\end{remark}

\subsection{Pre- and post processing}\label{sec:prepost}

In order to improve clustering, we may perform pre- or post-processing. For instance, in color image segmentation, we pre-process images with Gaussian blur.
Post-processing might be needed if the resulting segmentation produces a 
fairly large number of clusters. One can merge clusters or run {\it Identify 
clusters} with a lattice of larger scale.

\newpage
\section{Clustering of Two-dimensional Exemplary Datasets}\label{sec:2dexamples}

\begin{figure}
\begin{center}
\begin{tabular}{ccc|ccc}
Ground Truth & Mean-shift & DB Scan & Moved & Lattice & DI Clust \\
\hline
\includegraphics[width=.135\linewidth]{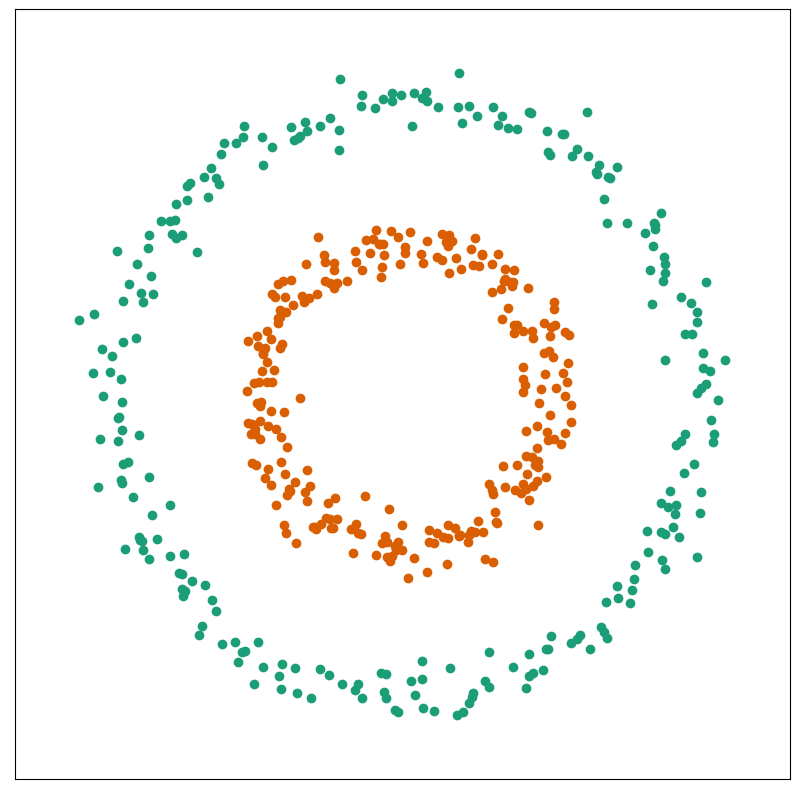}  &
\includegraphics[width=.135\linewidth]{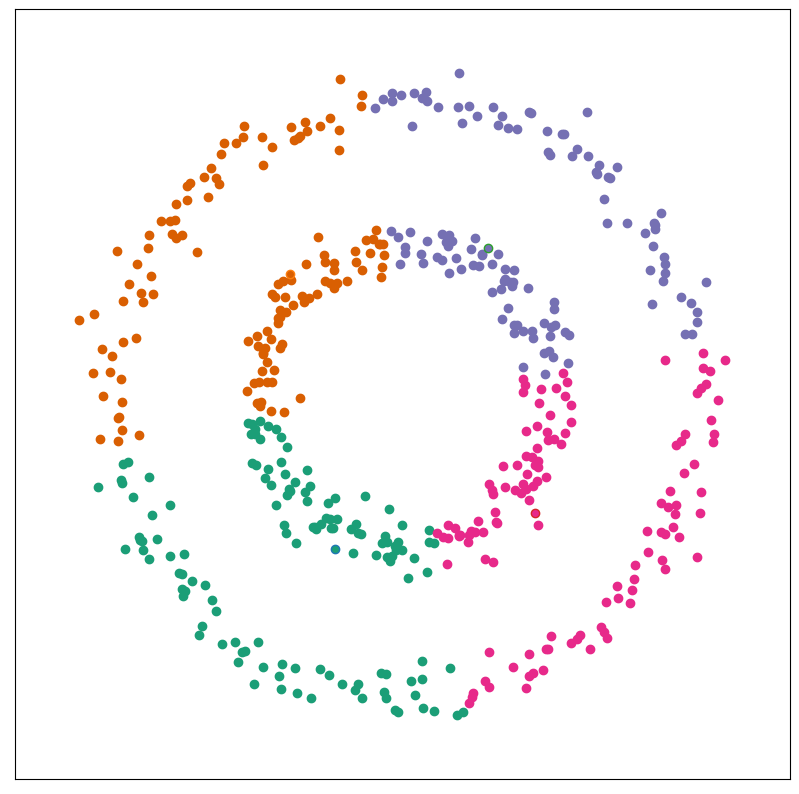}  &
\includegraphics[width=.135\linewidth]{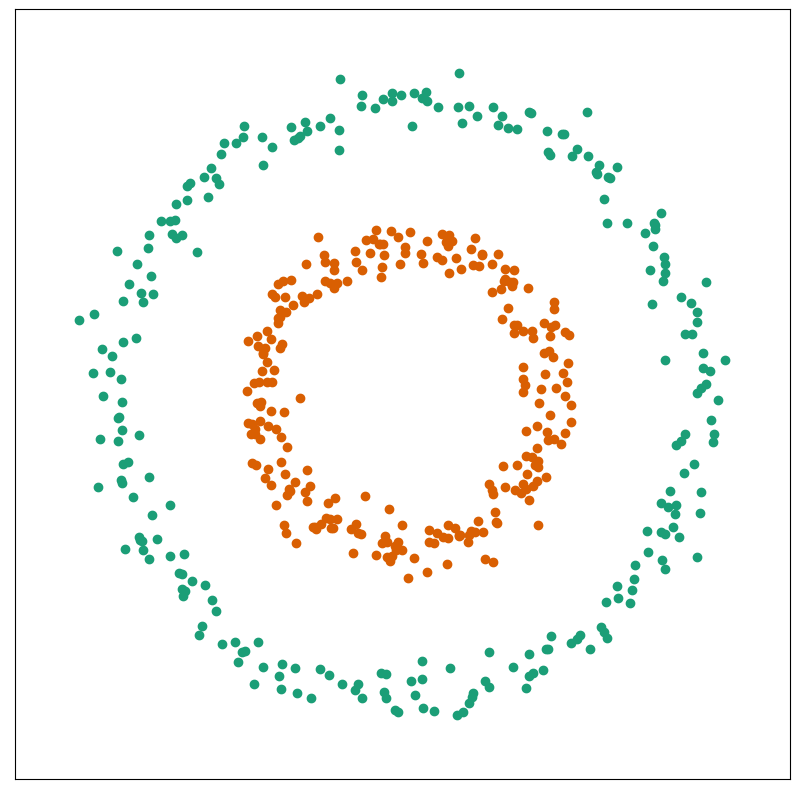}  &  
\includegraphics[width=.135\linewidth]{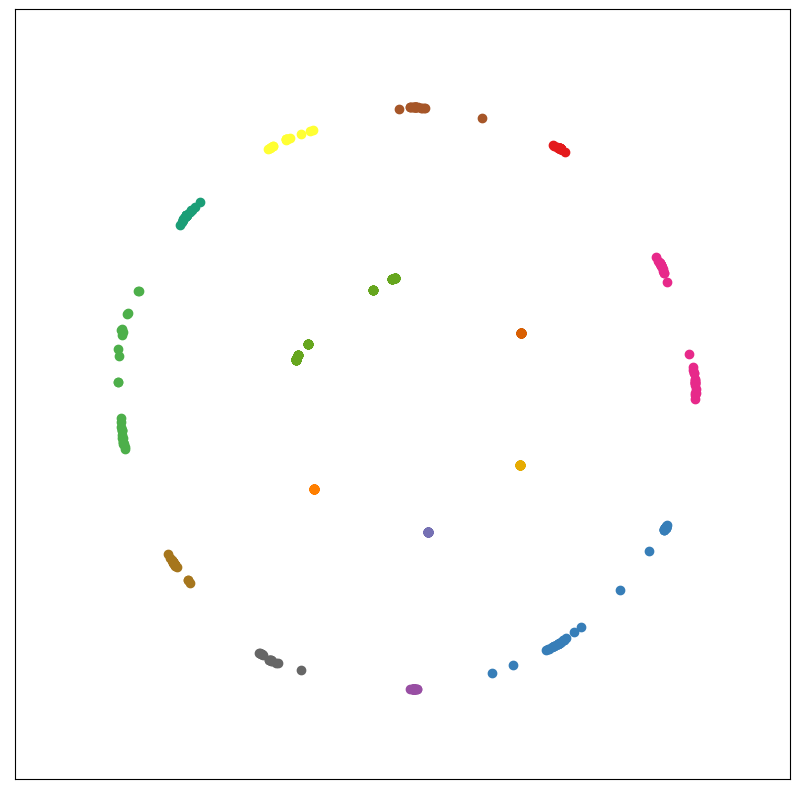}  &
\includegraphics[width=.135\linewidth]{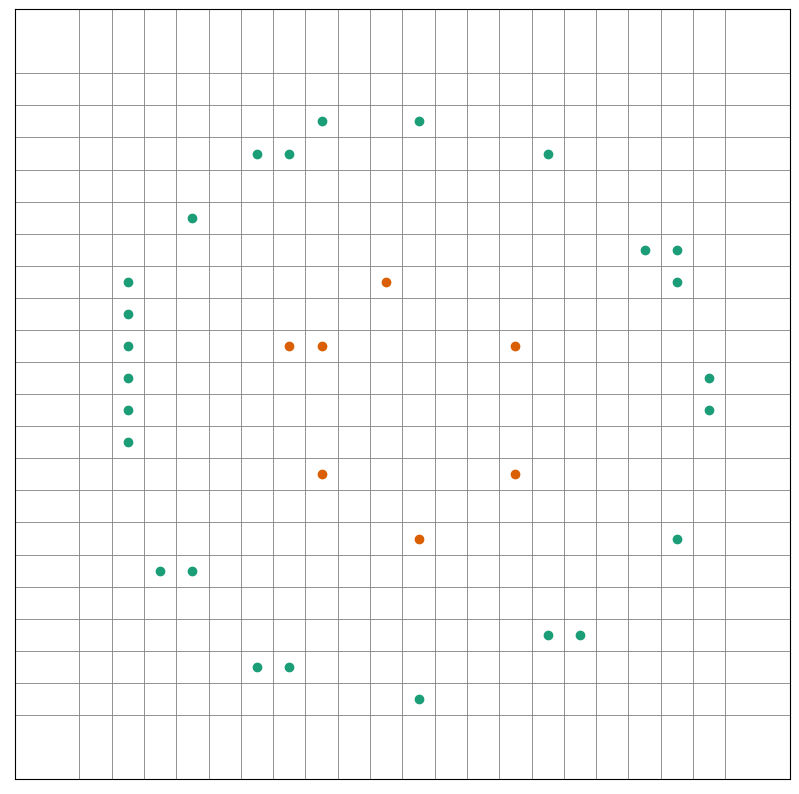}  &
\includegraphics[width=.135\linewidth]{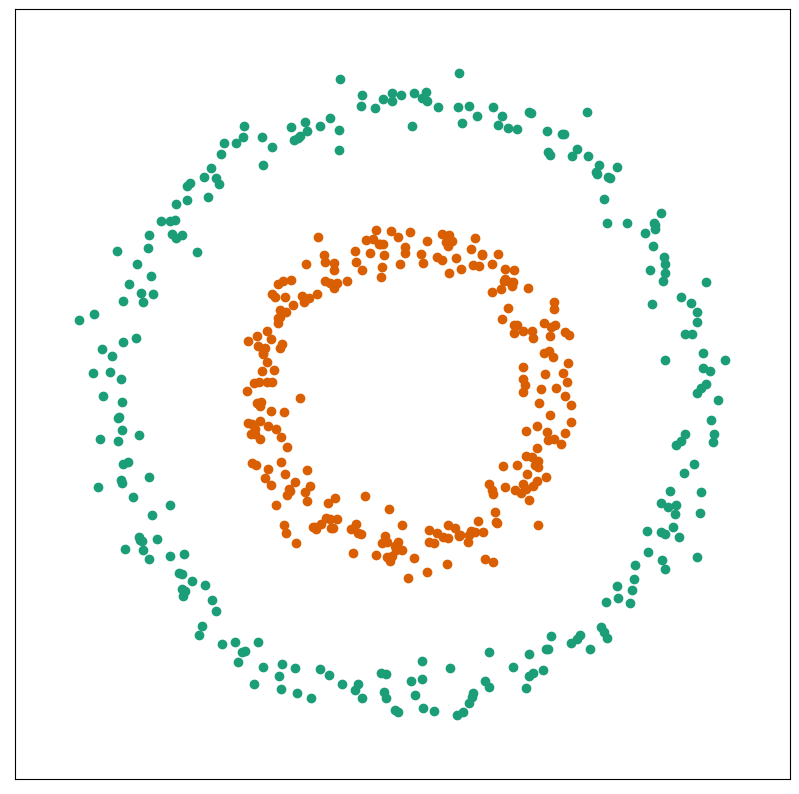}  \\
\includegraphics[width=.135\linewidth]{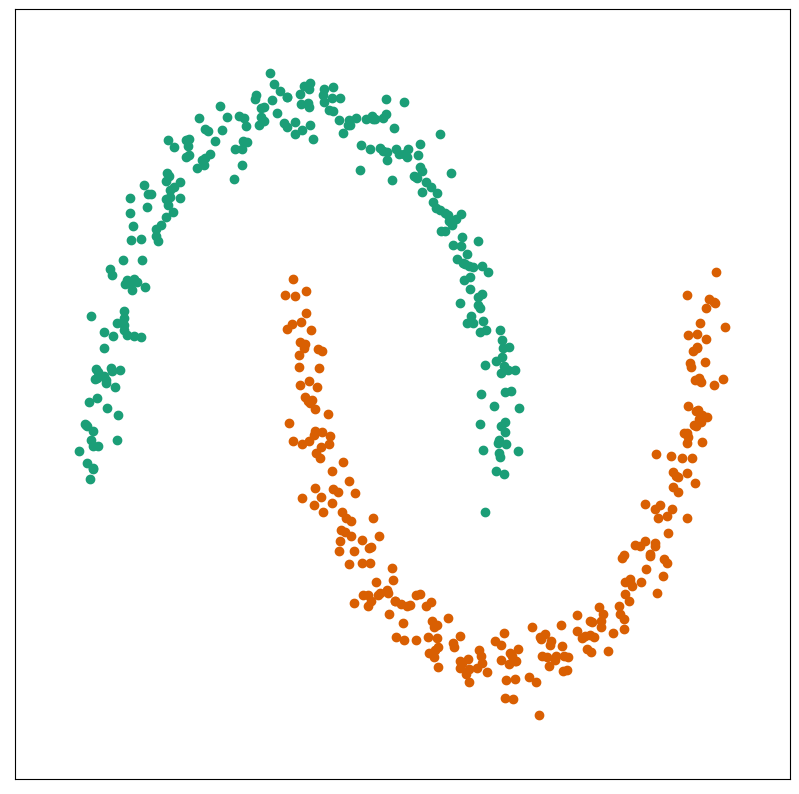}  &
\includegraphics[width=.135\linewidth]{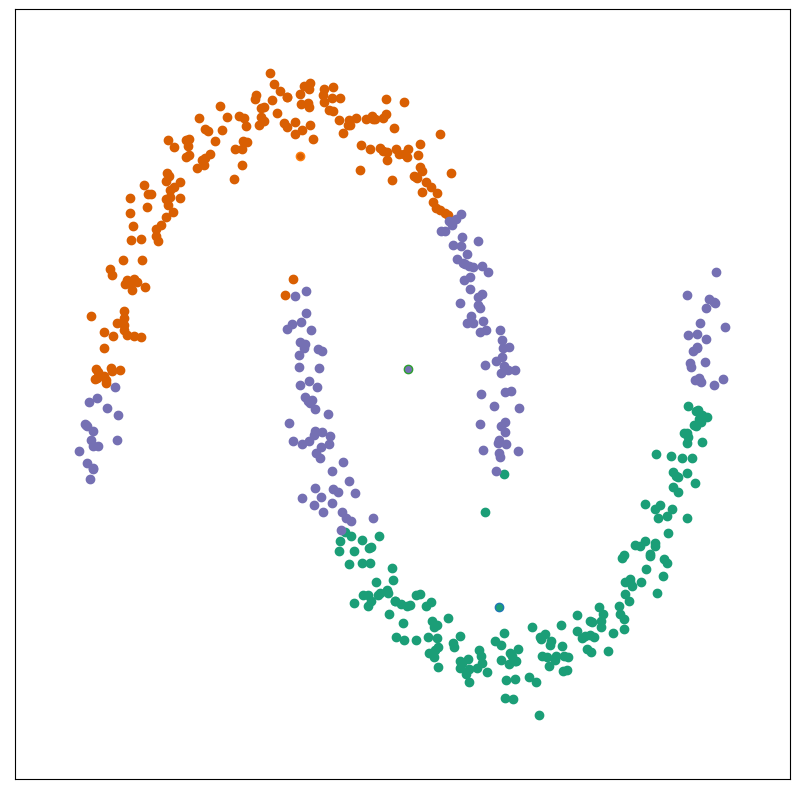}  &
\includegraphics[width=.135\linewidth]{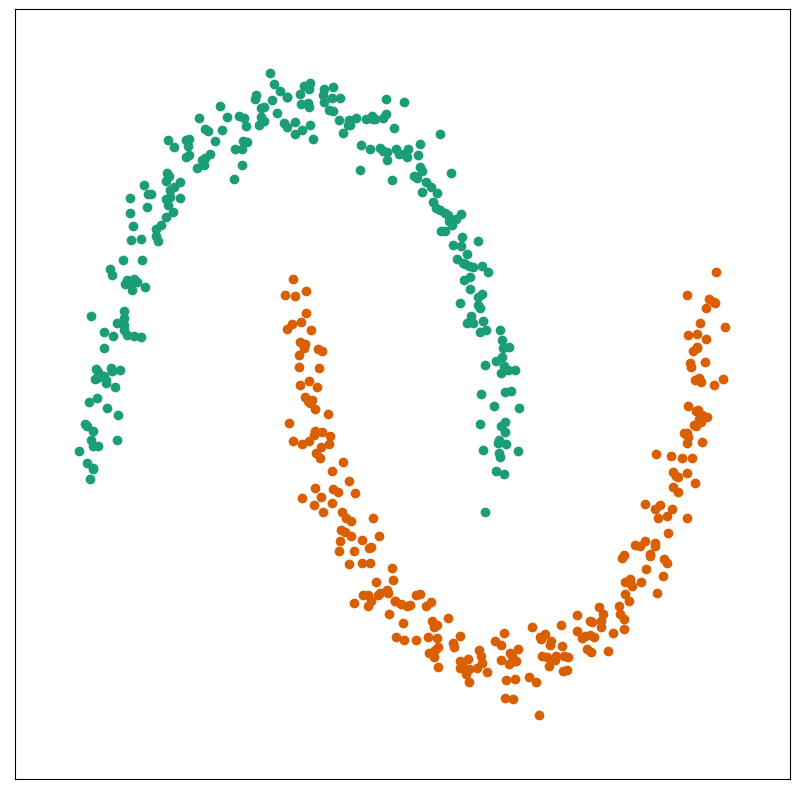}  &  
\includegraphics[width=.135\linewidth]{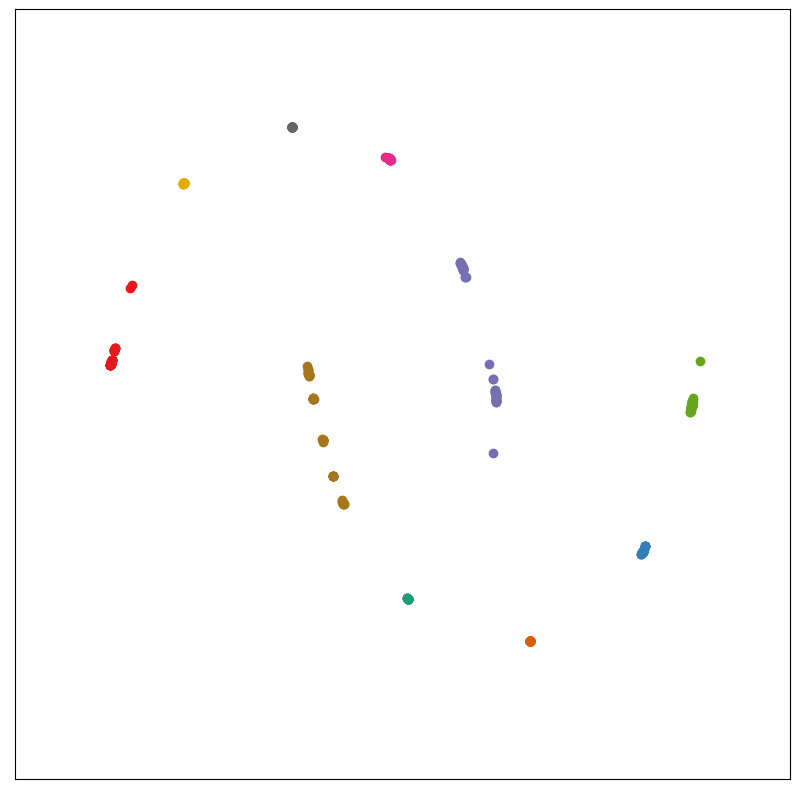}  &
\includegraphics[width=.135\linewidth]{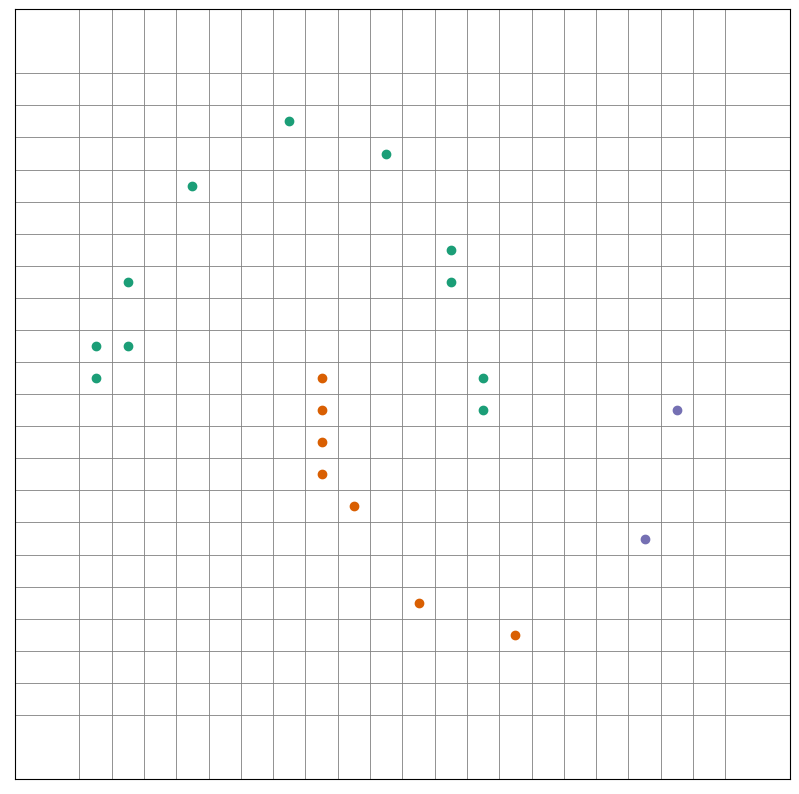}  &
\includegraphics[width=.135\linewidth]{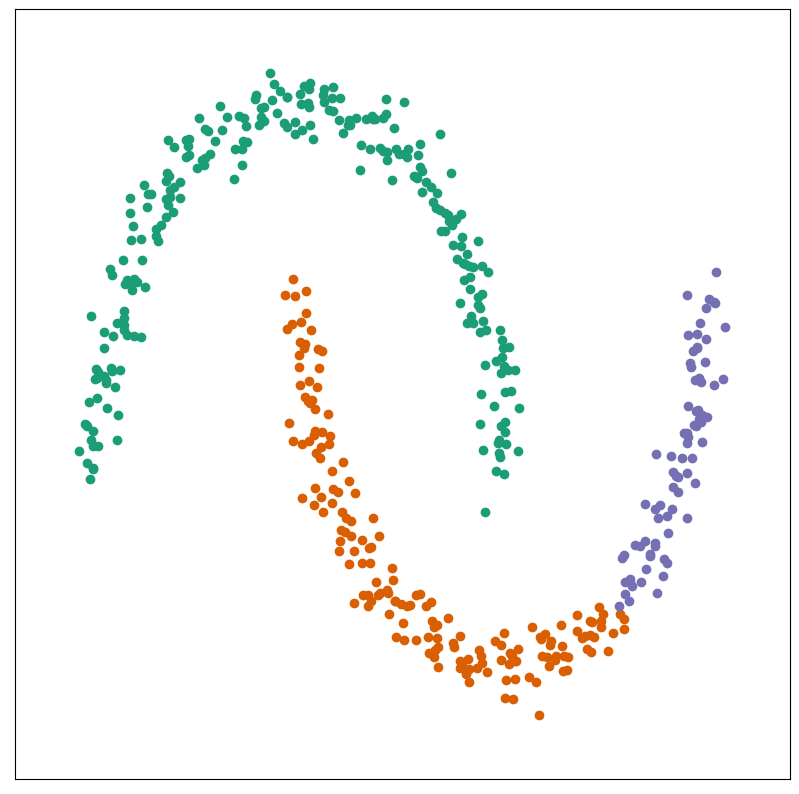}  \\
\includegraphics[width=.135\linewidth]{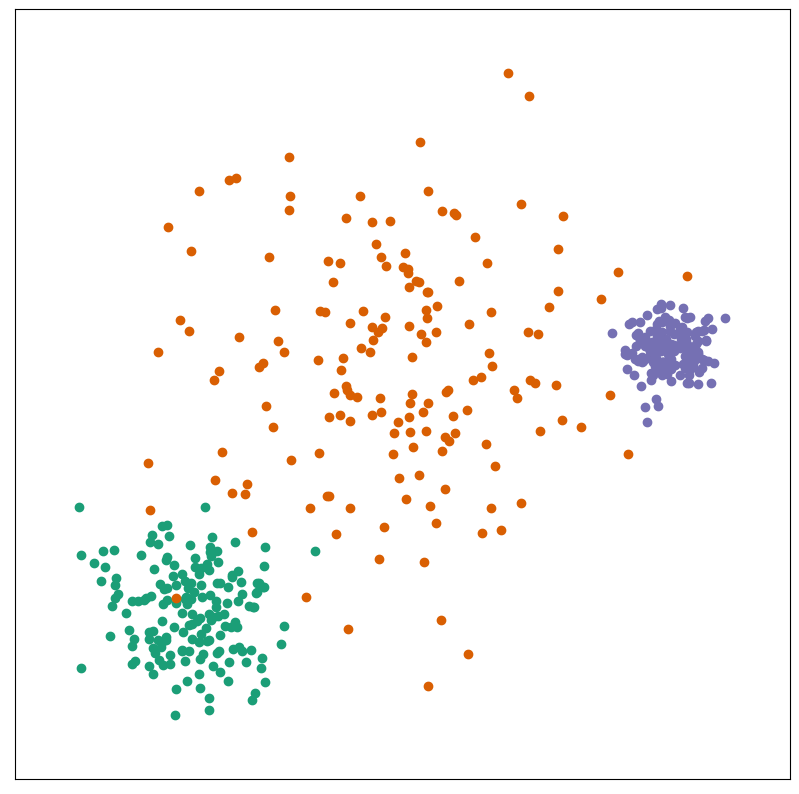}  &
\includegraphics[width=.135\linewidth]{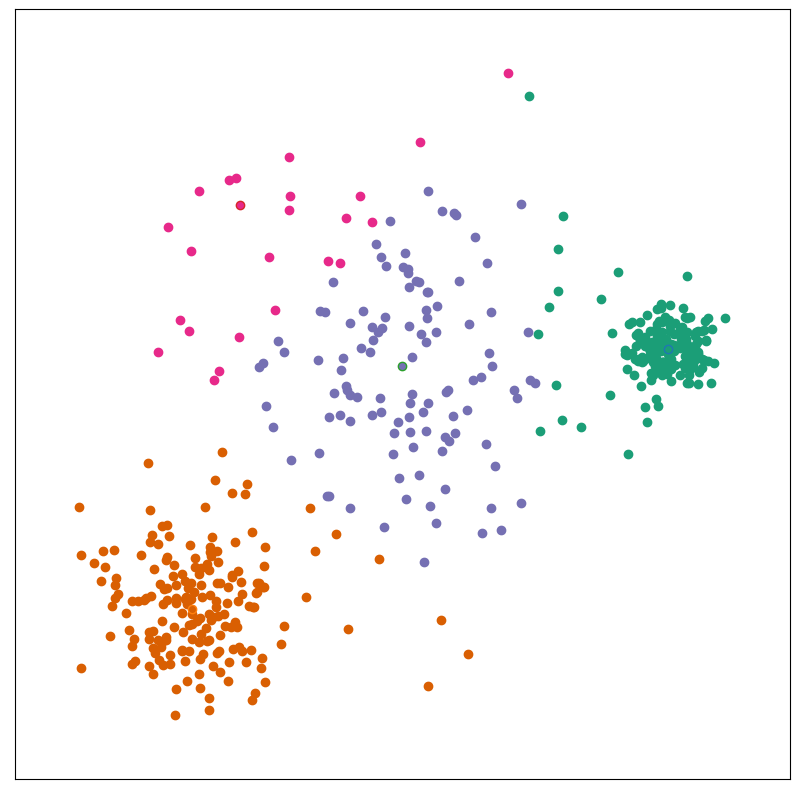}  &
\includegraphics[width=.135\linewidth]{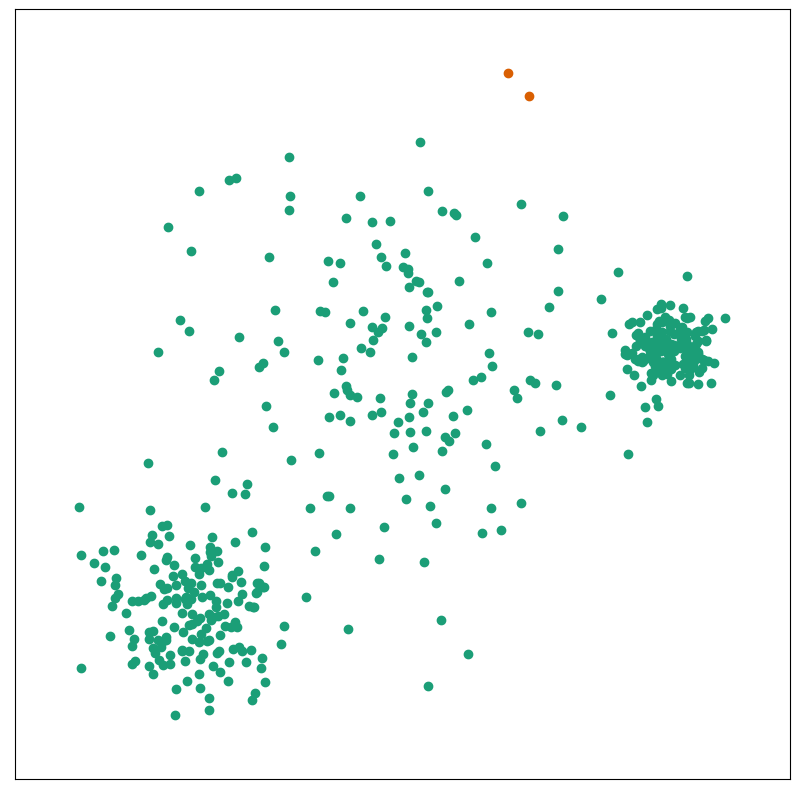}  &  
\includegraphics[width=.135\linewidth]{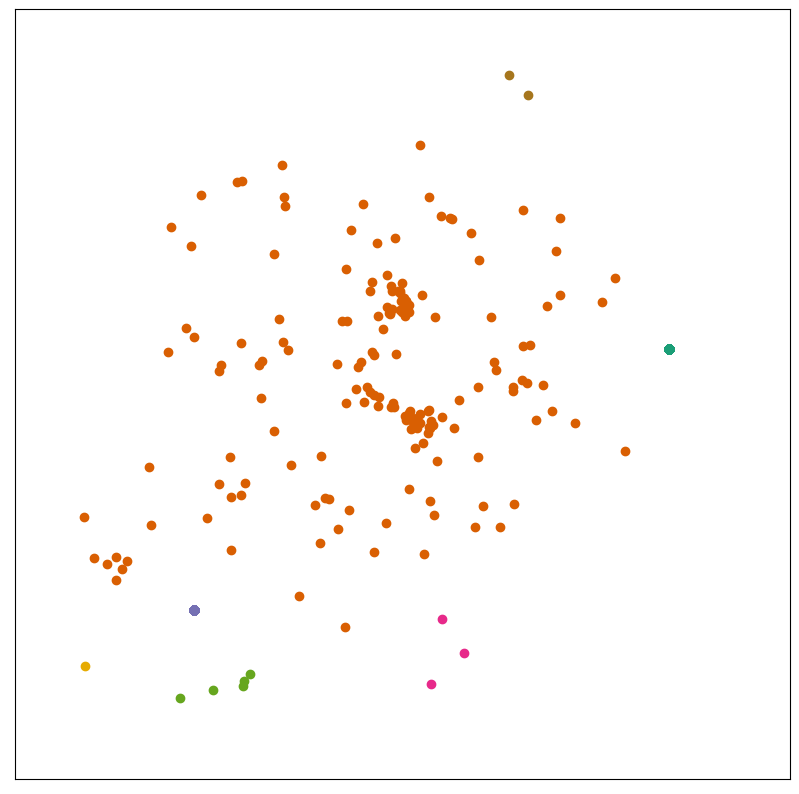}  &
\includegraphics[width=.135\linewidth]{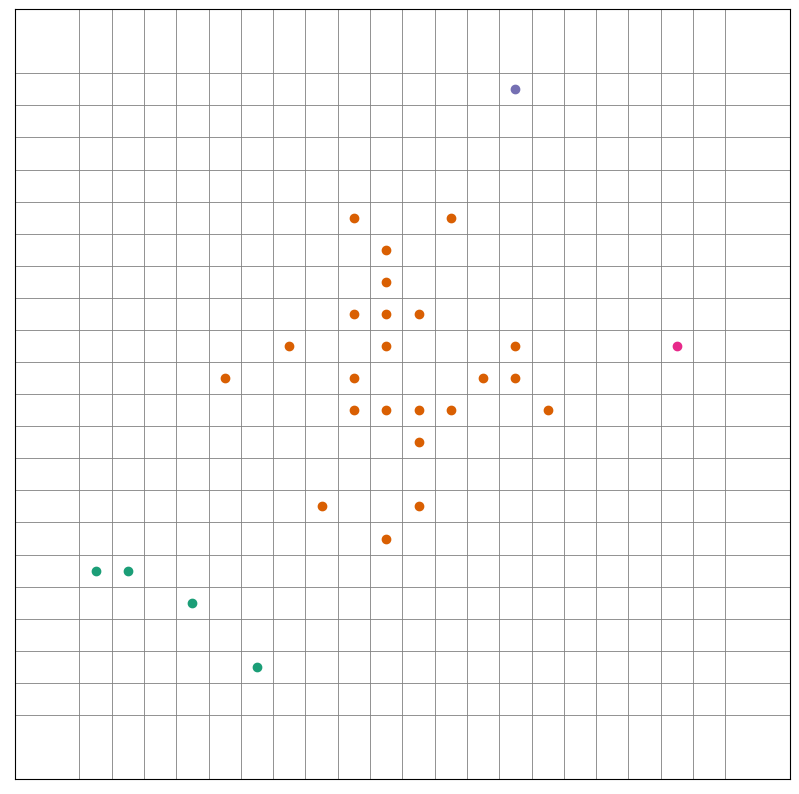}  &
\includegraphics[width=.135\linewidth]{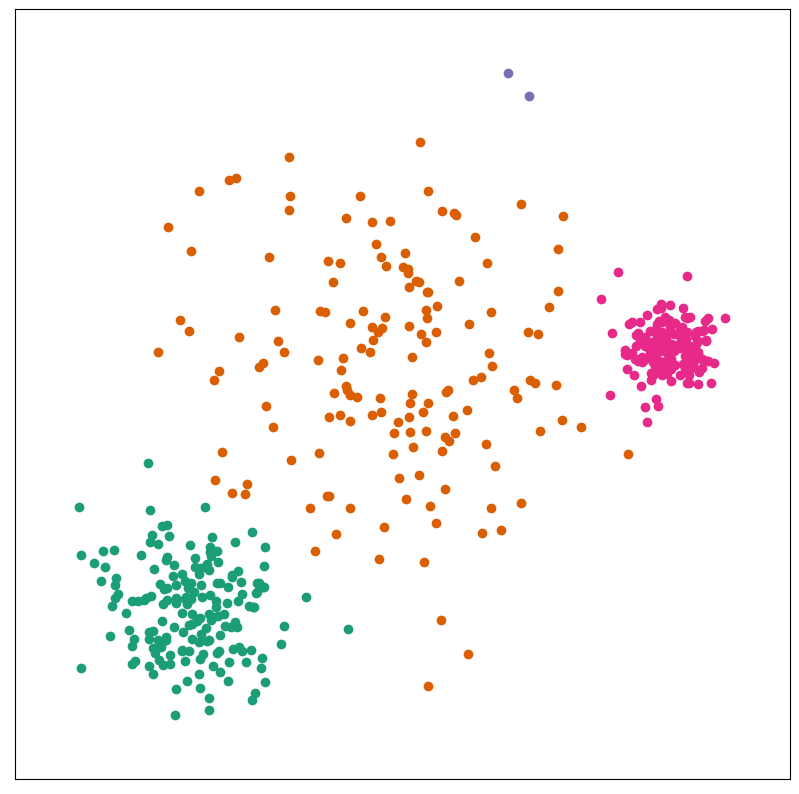}  \\
\includegraphics[width=.135\linewidth]{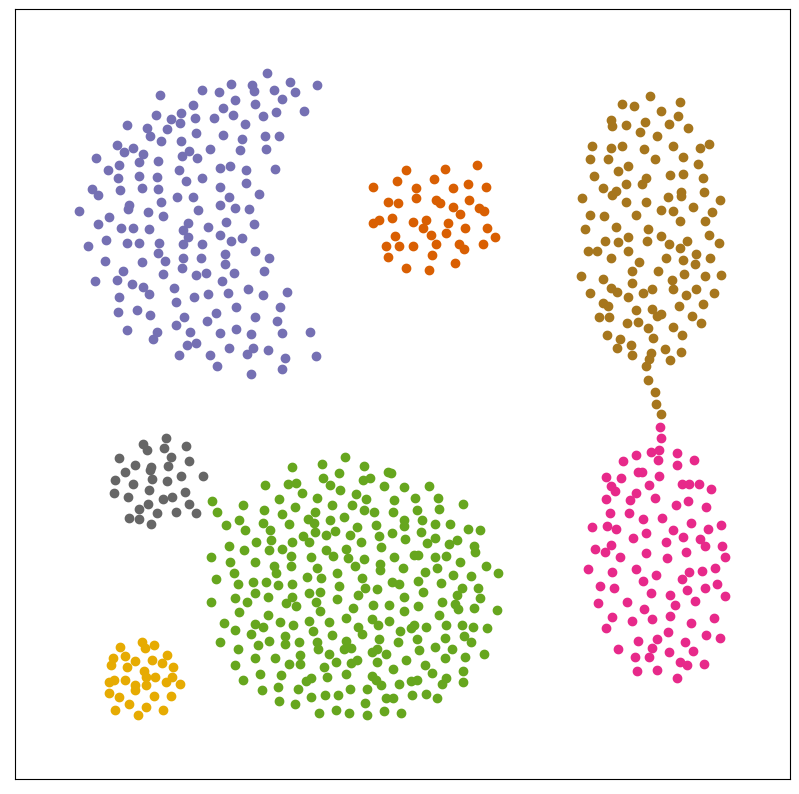}  &
\includegraphics[width=.135\linewidth]{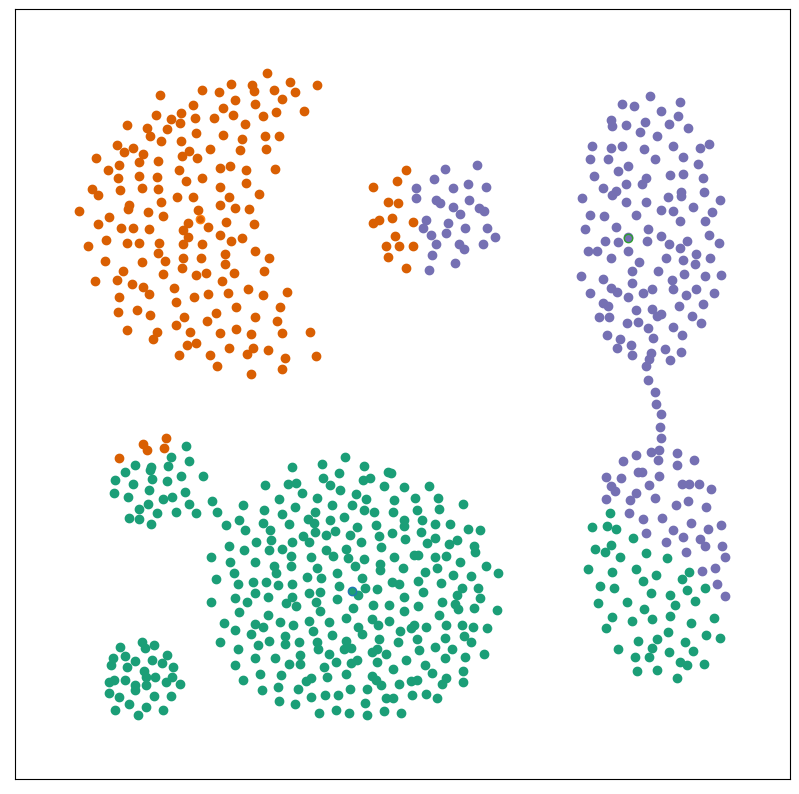}  &
\includegraphics[width=.135\linewidth]{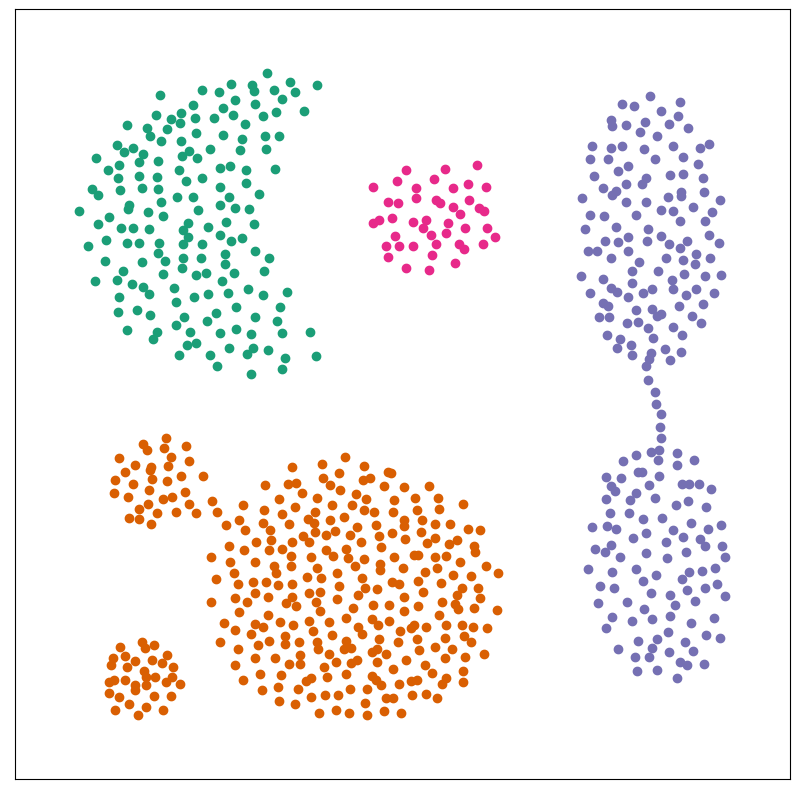}  &  
\includegraphics[width=.135\linewidth]{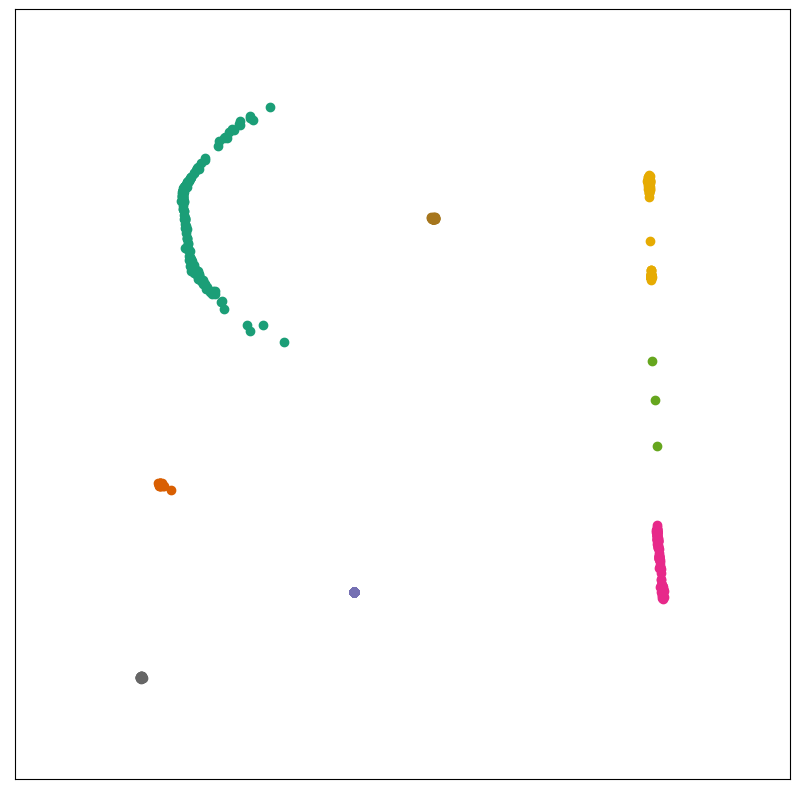}  &
\includegraphics[width=.135\linewidth]{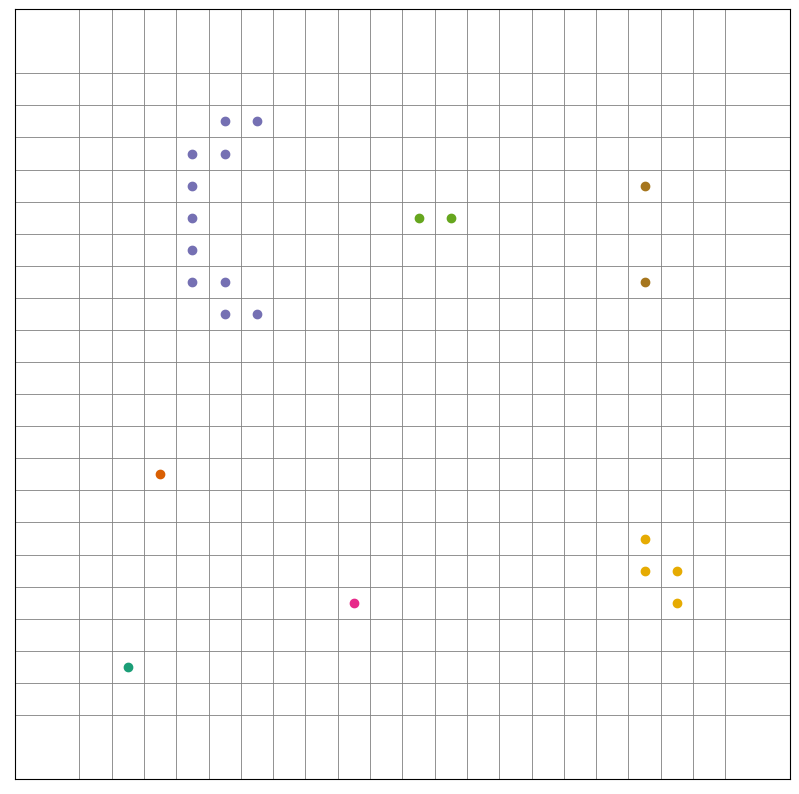}  &
\includegraphics[width=.135\linewidth]{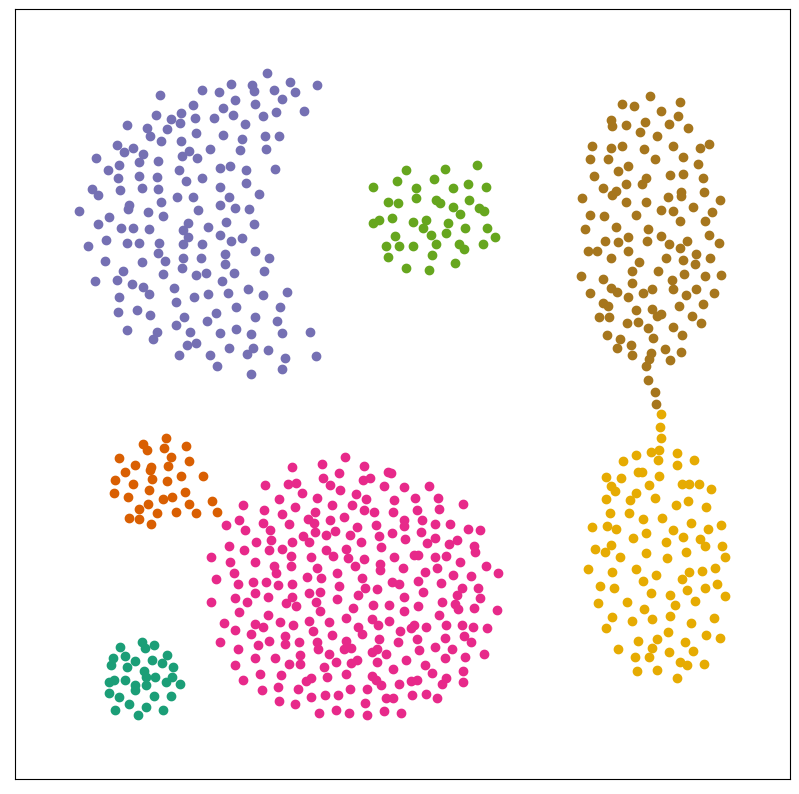}  \\
\includegraphics[width=.135\linewidth]{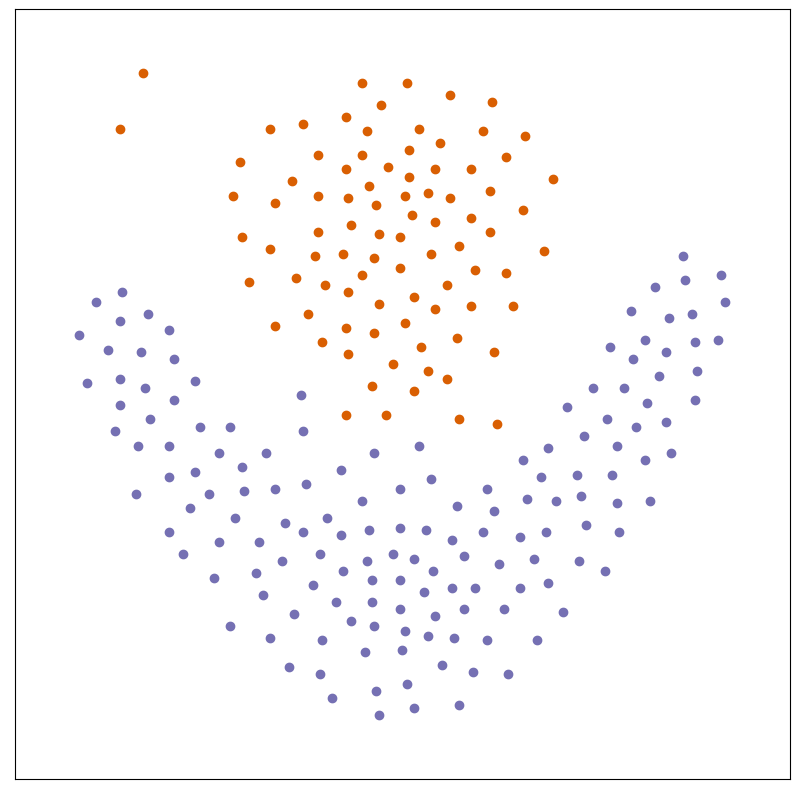}  &
\includegraphics[width=.135\linewidth]{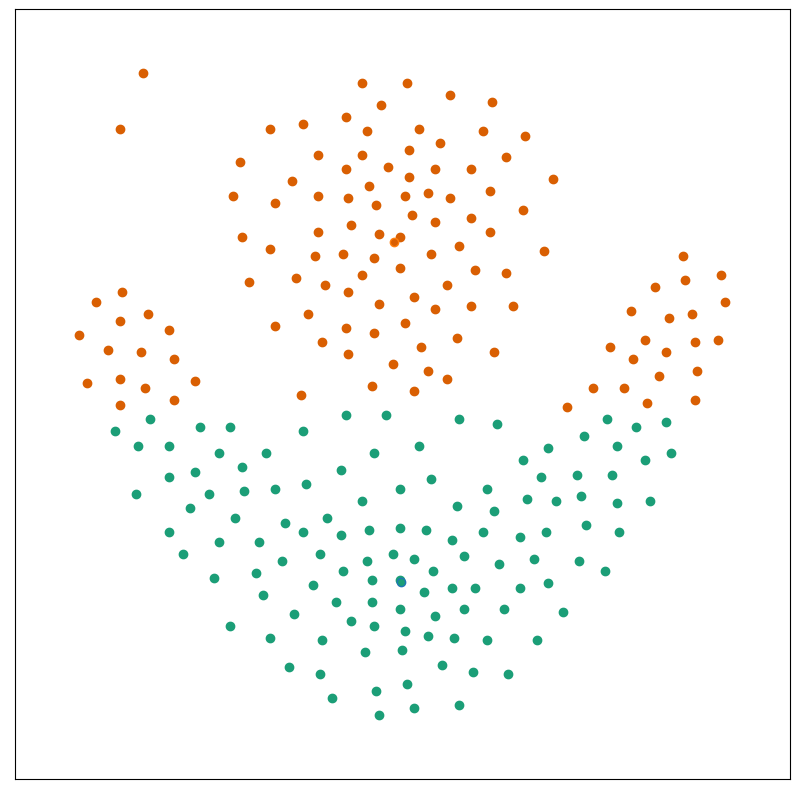}  &
\includegraphics[width=.135\linewidth]{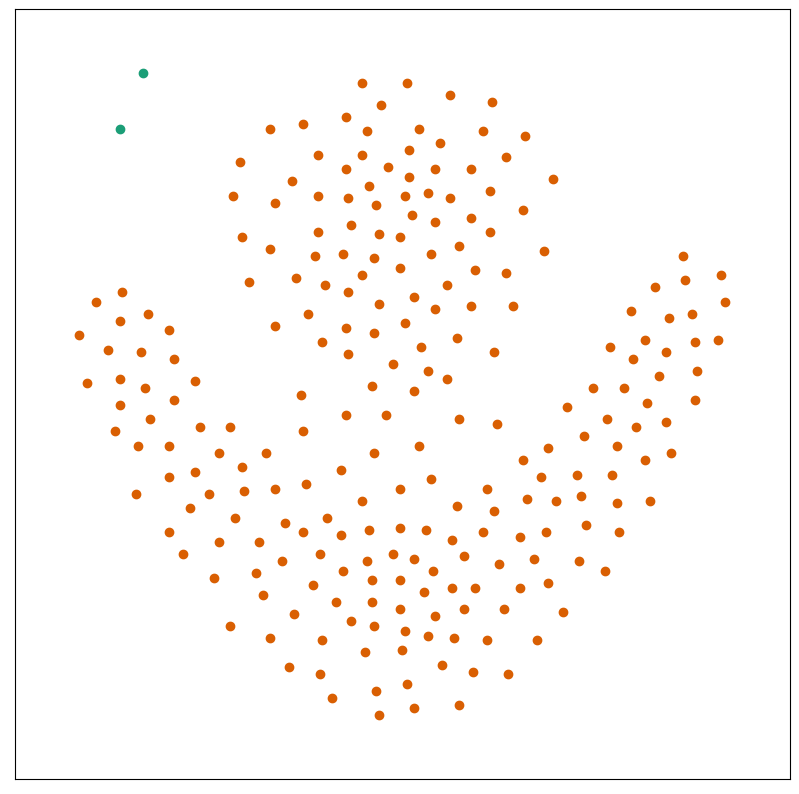}  &  
\includegraphics[width=.135\linewidth]{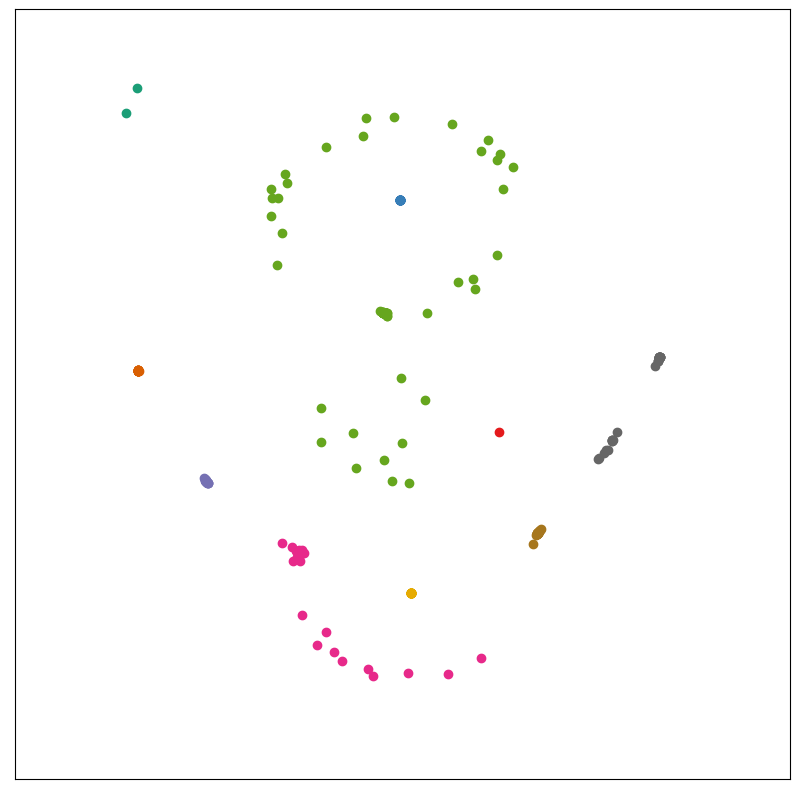}  &
\includegraphics[width=.135\linewidth]{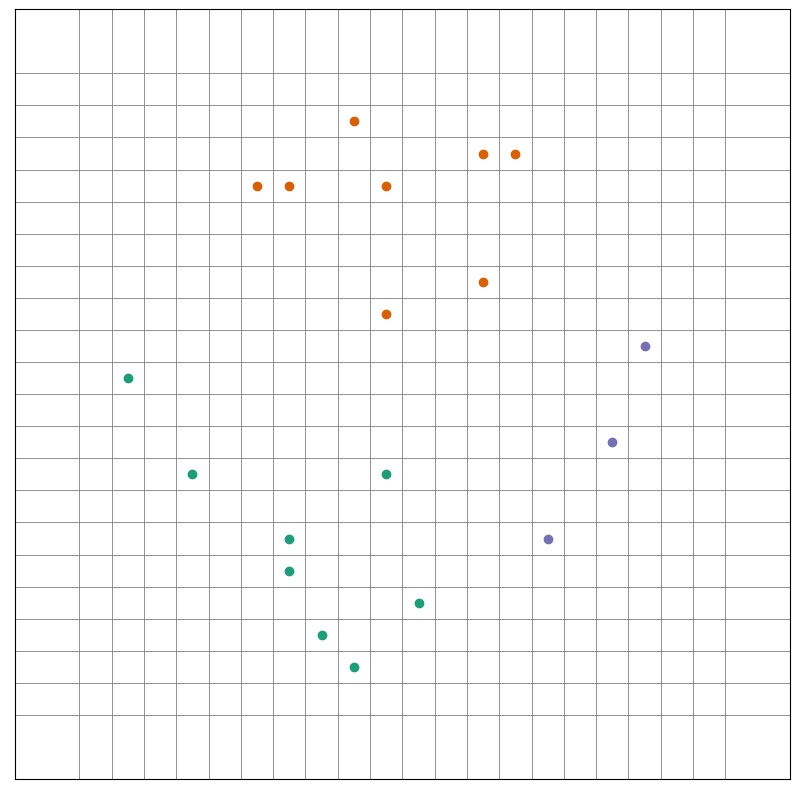}  &
\includegraphics[width=.135\linewidth]{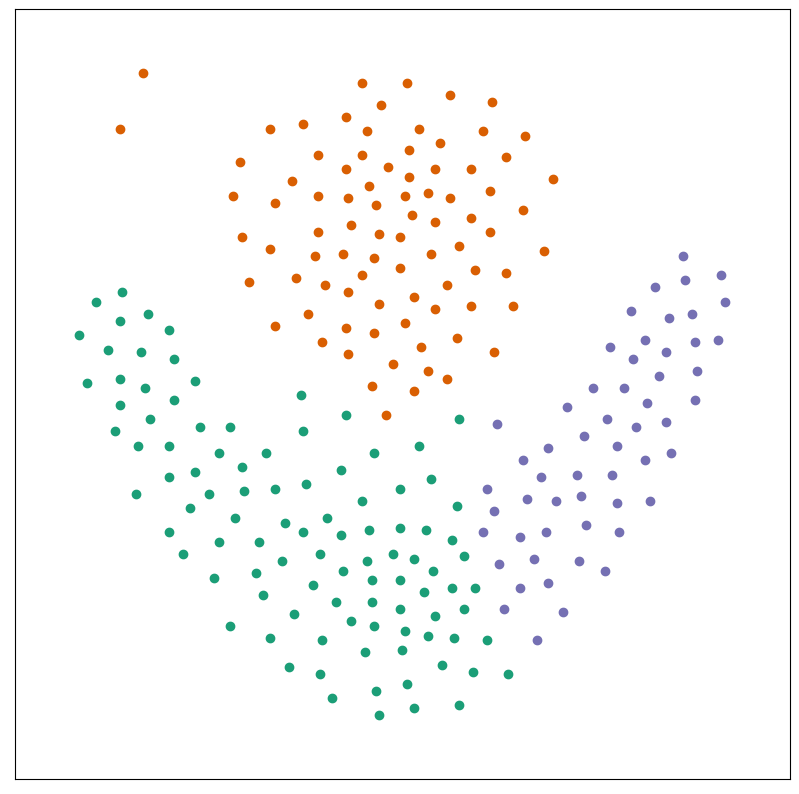}  \\
\end{tabular}
\end{center}
 \caption{   }
\label{fig:comparison}
\end{figure}

In this section, we present a comparison between our DI clustering and two 
classical unsupervised 
methods: mean-shift and DBSCAN. Each row of Fig. \ref{fig:comparison} presents an example of 
shape dataset, see~\cite{ClusteringDatasets} and references therein. In the first column we see 
the ground-truth clustering. In the second and third -- mean-shift and DBSCAN, respectively. 
The remaining columns are dedicated to the DI protocol. 

The parameters for the applied algorithms are as follows
$m=1$, $\delta=0.1$, $\epsilon = \frac{\delta}{2}=0.05$ and $n_{max}= 40$
DBSCAN is run with the parameters $\delta$ and $m$ that are the same as in DI 
protocol. Mean-shift uses default values and bandwidth estimation, 
see~\cite{scikit-learn} for details.

In the forth column of Fig.~\ref{fig:comparison} pixels/agent are in position 
after the data was advanced $n_0$ steps in time. In Algorithm 1, it corresponds 
to the situation at the end of the loop. Then the fifth column presents the 
situation after building the $\epsilon$-lattice and assigning clusters within 
the lattice. Finally, column six represents the clustering applied to the 
original datasets. 

Note that the parameters used in Fig.~\ref{fig:comparison} were not 
particularly chosen to suit the DI protocol at the expense of other algorithms. 
The conclusion related to the accuracy of the algorithm (and disregarding 
numerical complexity) that arises upon examination of Fig.~\ref{fig:comparison} 
and similar pictures is as follows:

\begin{itemize}
    \item The DI protocol, similar to the DBSCAN algorithm and contrary to the 
    mean-shift, is suited to cluster data with arbitrary shapes.
    \item The DI protocol tends to split clusters, which in many cases prove 
    useful, e.g. example 3 and 4 and in other times can be detrimental, e.g. 
    example 2 and 5. This is further visualized and discussed in Fig. 
    \ref{fig2} in the next section.
    \item The advantage of the DI protocol lies in simplification of data and 
    not necessarily clustering itself. Particularly, columns 4 and 5 present a 
    simplification of the original ground truth in the first column. The 
    general shape of the data is retained but the distributions are less 
    complex. This may prove useful as a method to reduce the dimension of the 
    data.
\end{itemize}

\section{Application to Color Image Segmentation}\label{sec:appl}

In order to further illustrate our method, we apply Algorithm \ref{alg} to  
selected pictures from Berkeley Segmentation Dataset (BSD) 
\cite{MartinFTM01}.
Size of the images is $481$ times $321$ what results in $N=154401$ pixels.

The position $x_i(t)\!\in\!\R^5$  represents the $i$th pixel in a 
5D feature space, where the modality is realised by combining 2D spatial positions of pixels in an image and their 3D representations in a color 
space ({\it e.g.} in $RGB$). In the feature space we define the combined 
distance $d(i,j)$ between pixels $i$ and $j$, the spatial distance 
$d_s(i,j)$ and color distance $d_c(i,j)$ as follows:
\begin{align*}
d_s(i,j)^2 &= (y^1_i-y^1_j)^2+(y^2_i-y^2_j)^2,\\
d_c(i,j)^2 &= (r_i-r_j)^2+(g_i-g_j)^2 + (b_i -b_j)^2,\\
d(i,j) &= \sqrt{d_s(i,j)^2+d_c(i,j)^2},
\end{align*}
where $y^1$ and $y^2$ are spatial coordinates and $r$, $g$, $b$ are color values in 
the $RGB$ color space.  All variables are normalised to $[0,1]$, therefore we 
work in a 5D unit cube.

\begin{figure}[h!]  
  \begin{picture}(0,385)
    \put(0,190){\includegraphics[width=0.5\textwidth]{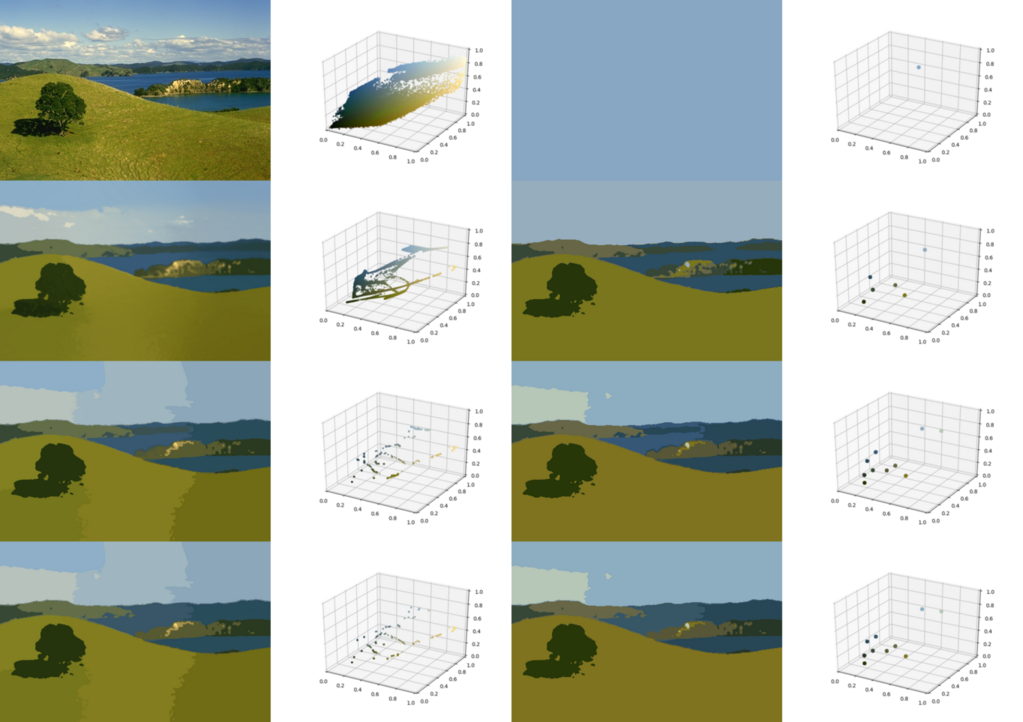}}
    \put(0,0){\includegraphics[width=0.5\textwidth]{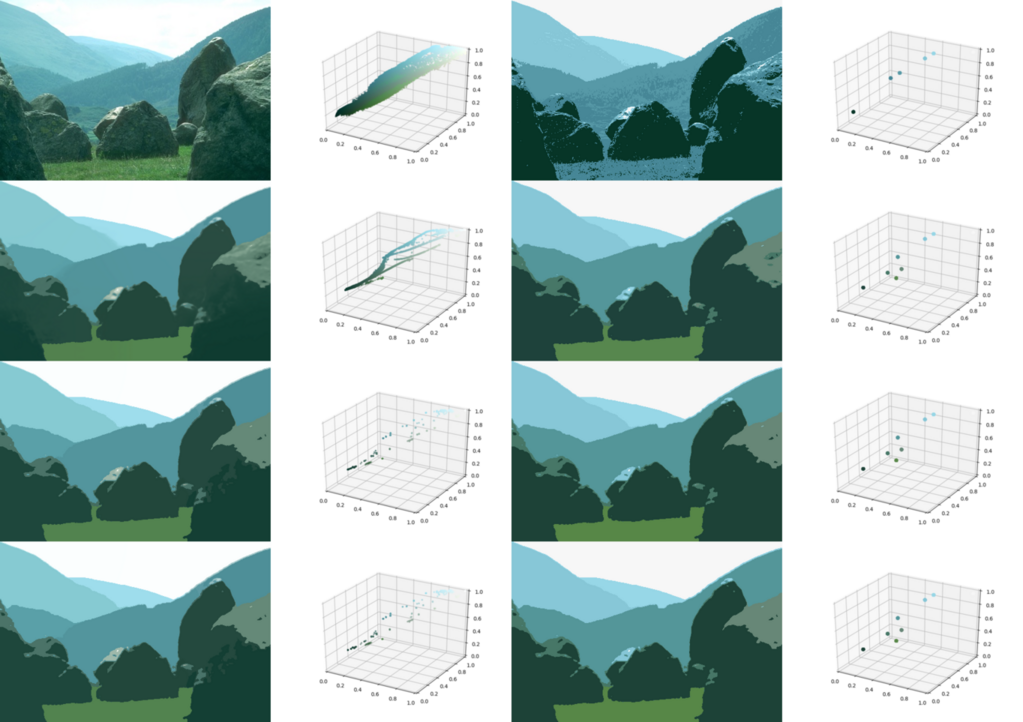}}
    \put(0,375){A)}
    \put(70,375){B)}
    \put(130,375){C)}
    \put(200,375){D)}
  \end{picture}
  \caption{Illustration of segmentation process, for parameters $\delta = 
  0.15$, $m=40$ and $n_{max} = 0,3,7,10$.}
  \label{fig2}
\end{figure}
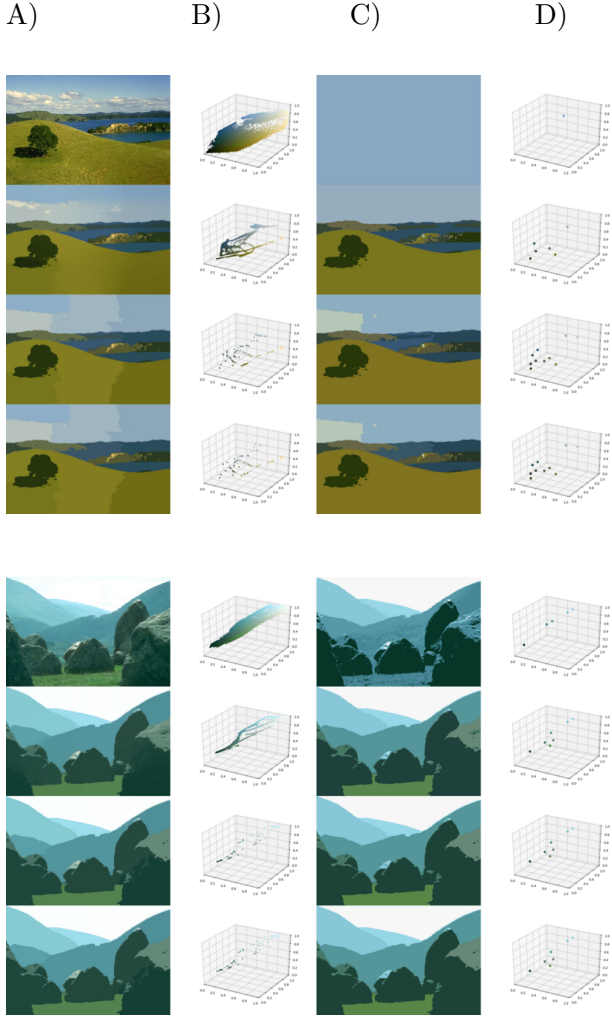

Segmentation follows Algorithm~\ref{alg}. We apply Gaussian blur as 
pre-processing and for post-processing we assign colors and outliers 
as described in Section \ref{sec:ass}. This process is illustrated in 
Figure~\ref{fig2}.

Column A of Figure~\ref{fig2} consists of $4$ images with original positions of 
the pixels and colors inherited from $0$, $3$, $7$ and $10$ iteration in each 
row, respectively. For instance,
the image in column A and row 2 presents the 5D pixels with combined 
coordinates $x^A_i(3)=(y_i^1(0),y_i^2(0),r_i(3),g_i(3),b_i(3))$.
Column B represents the projection of each pixel $x^A_i$ onto the 3D $RGB$ 
space, {\it e.g.} for $x^A_i(3)$ we have $(r_i(3),g_i(3),b_i(3))$ in the image 
in column B and row 2.
Column C consists of images with original positions and colors obtained through 
the application of the cluster identification steps of Algorithm~\ref{alg} 
(c.f. Sections \ref{sec:dpl} and \ref{sec:ass}) with $n_{max} = 0,3,7,10$, 
respectively. Column D represents positions of clusters in the $RGB$ space.

Observe the interesting case of the image in Column C and row 1. Since no 
iterations of the loop in Algorithm~\ref{alg} was performed, the pixels are not 
sufficiently densely packed and only one cluster, say ${\mathcal A}_1$, is 
identified by the procedure in Section \ref{sec:dpl}. Then, according to 
Section \ref{sec:ass}, the color of each pixel in ${\mathcal A}_1$ is changed 
to the average color of all pixels in ${\mathcal A}_1$, which happens to be 
blue. All remaining pixels outside of ${\mathcal A}_1$ are outliers and are 
then assigned the same blue color.

\begin{figure}[h!]
  \includegraphics[width=0.15\textwidth]{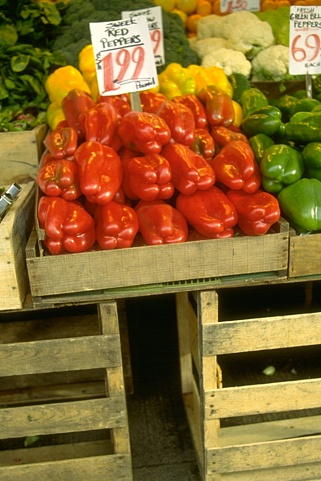}
  \caption{Peppers, input picture for parameter discussion.}
  \label{fig:pepper}
\end{figure}

In order to show the parameters' influence, we apply Algorithm~\ref{alg} 
with various $m$ and $\delta$ to a picture of peppers, 
Figure~\ref{fig:pepper}.

We pick the interaction range $\delta \in\{0.1,0.2\}$ and $\epsilon = \frac{\delta}{4}$. This choice is 
connected to a number of pixels $N$ and the dimension $d$,  $\delta \approx 
\sqrt[5]{\frac{1}{N}} \approx 0.1$. This corresponds to the length of an edge 
of a $5-$dimensional cube with volume $1/N$ (average volume for one pixel).

The parameter $m$ is set to be equivalent to the minimal 
cluster density being $\eta$ times greater than an average pixel density, 
$m = \frac{8}{15}\pi^2\eta N \delta^5$, with $\eta\in\{1,5\}$. 

\subsection{Parameters}\label{sec:params}

\begin{figure}[ht!]  
  \begin{picture}(0,300)
  \put(30,0){\includegraphics[width=0.5\textwidth]{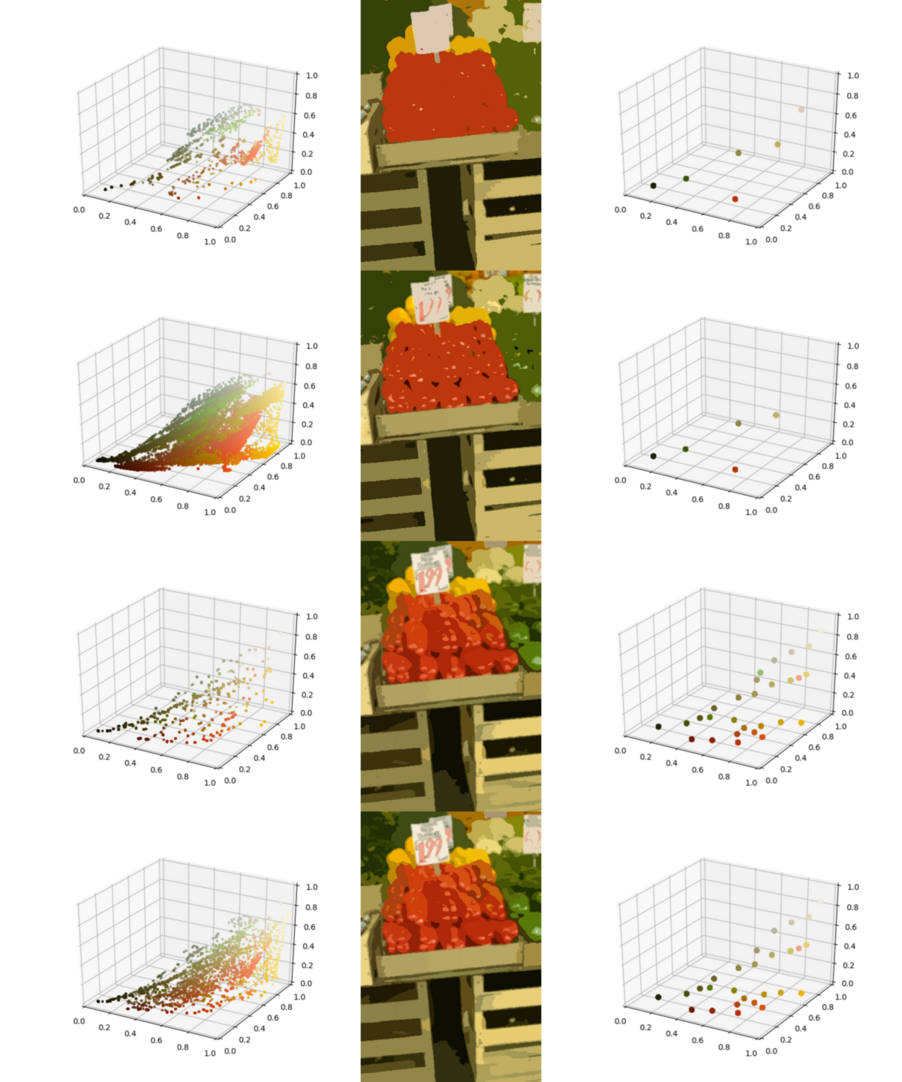}}
  \put(0,290){B)}
  \put(120,290){C)}
  \put(180,290){D)}
  \put(5,275){$\delta = 0.2, m=260, \eta=1$}
  \put(5,205){$\delta = 0.2, m=1300, \eta=5$}
  \put(5,135){$\delta = 0.1, m=8, \eta=1$}
  \put(5,65){$\delta = 0.1,  m=40, \eta=40$}
\end{picture}
  \caption{Illustration of the influence of parameters.}
  \label{fig:params}
\end{figure}
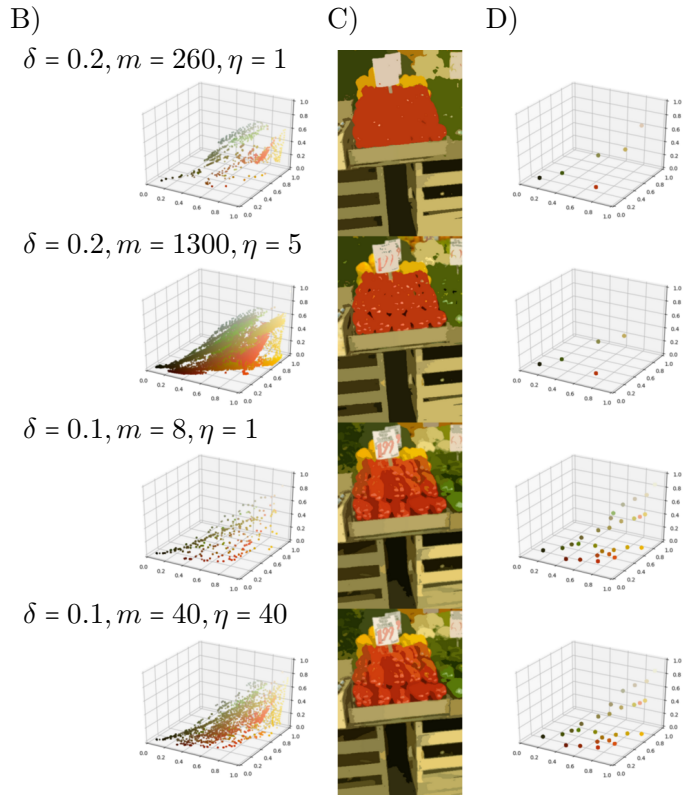

Figure~\ref{fig:params}, presents: fully segmented image (Column C) with $n_{max}=10$, and two
projections onto the 3D $RGB$ space (Columns B and D), as described above for Figure~\ref{fig1}.
We observe that the number of clusters is directly connected to the interaction 
range $\delta$, but seems mostly independent of parameter $m$. 
Instead, parameter $m$, drives the dynamics of agents, c.f. Column B, and thus -- the 
final cluster assignment. 
Indeed, comparing rows 1 and 2 in Fig. \ref{fig:params}, we observe that the 
clusters in Column D are almost the same, but the segmented images in Column C differ.

Note, that in extreme cases, for large $\delta$ and small $m$, one will obtain 
only one cluster. On the other hand, with small $\delta$ and large $m$ each 
point will be placed into its cluster, like in the input image.

Unsupervised color image segmentation is an image processing task without 
a unique result. The BSD provides several human-performed segmentations for each picture. 
As pointed out in \cite{Zhang2008} one cannot guarantee that any 
manually-generated segmentation image is better than another. This is 
especially the case with nature images. 
Thus, the dependence on parameters can be seen as an advantage of the 
presented method, allowing to encapsulate different results of color image segmentation, similar to the manually-generated cases.

Note that the provided parameters are closely related to the average density of 
pixels. In practice, taking $\delta$ and $m$ from small ranges of values leads 
to relatively small differences in the resulting segmented pictures, which 
corresponds to the slight 
variation between human-generated segmentations. Thus, restriction of the set of admissible parameters to values related to average density of the pixels, while ensuring that bounds established in Section \ref{sec:complex} are satisfied, makes finding the proper values of parameters manageable. This is further showcased by Fig. \ref{fig:segresults} below, where multiple different pictures are segmented using the presented method with the same values of parameters leading to mostly reasonable results.

\subsection{Segmentation Results on BSDS500}

\begin{figure}[h!]
 \begin{center}
 \includegraphics[width=0.24\textwidth]{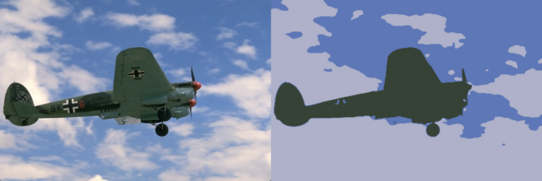}
 \includegraphics[width=0.24\textwidth]{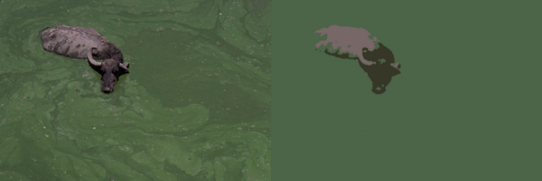}
 \includegraphics[width=0.24\textwidth]{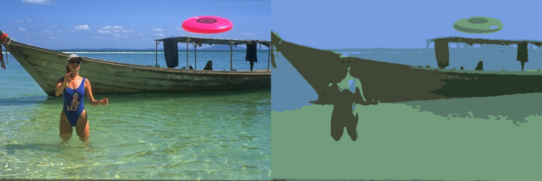}
 \includegraphics[width=0.24\textwidth]{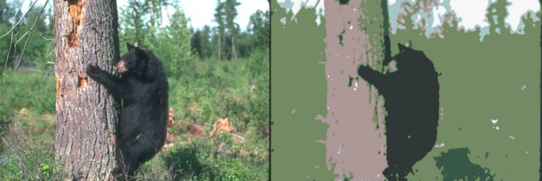}
 \includegraphics[width=0.24\textwidth]{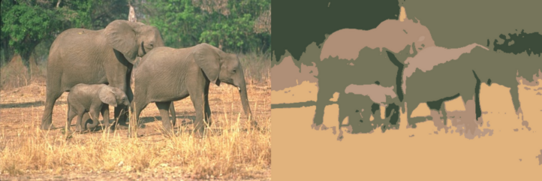}
 \includegraphics[width=0.24\textwidth]{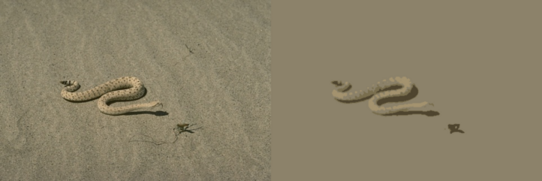}
 \includegraphics[width=0.24\textwidth]{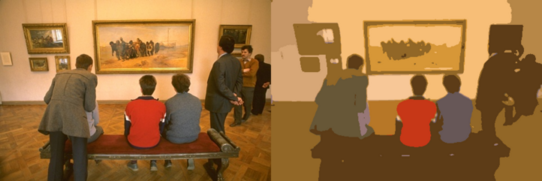}
 \includegraphics[width=0.24\textwidth]{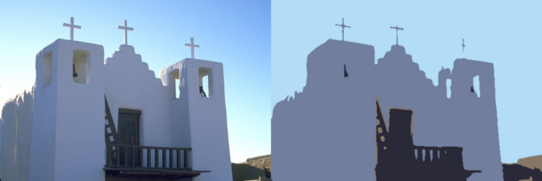}
 \includegraphics[width=0.24\textwidth]{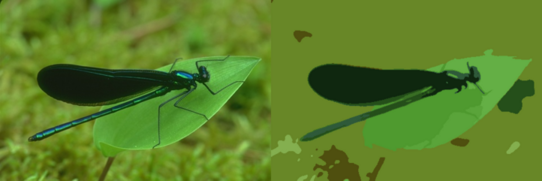}
 \includegraphics[width=0.24\textwidth]{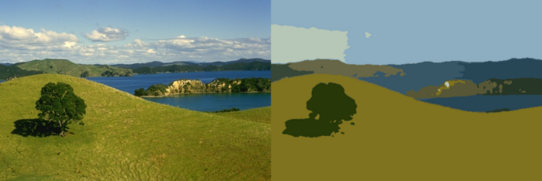}
 \includegraphics[width=0.24\textwidth]{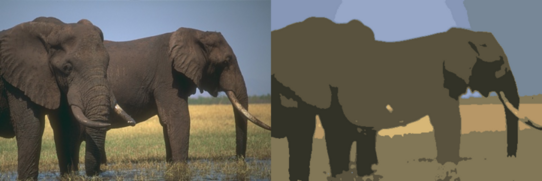}
 \includegraphics[width=0.24\textwidth]{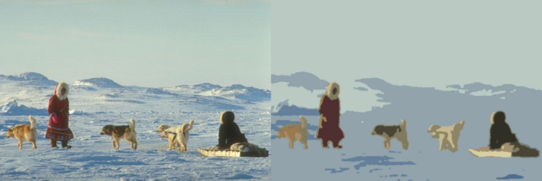}
 \includegraphics[width=0.157\textwidth]{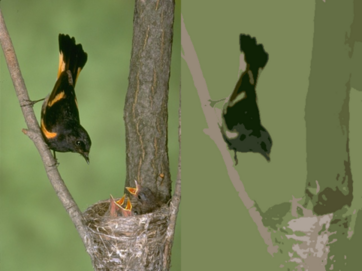}
 \includegraphics[width=0.157\textwidth]{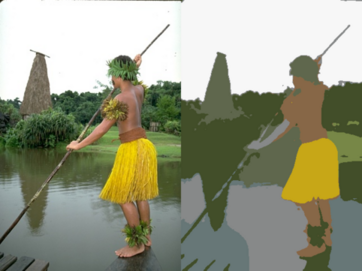}
 \includegraphics[width=0.157\textwidth]{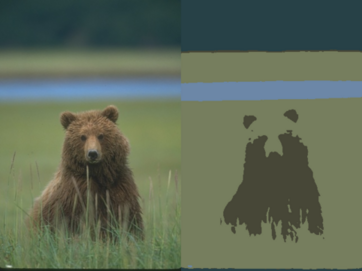}
 \includegraphics[width=0.157\textwidth]{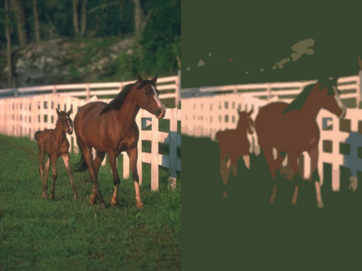}
 \includegraphics[width=0.157\textwidth]{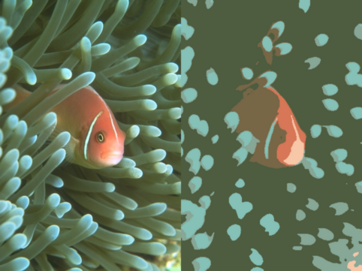}
 \includegraphics[width=0.157\textwidth]{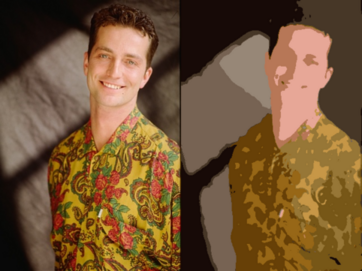}
\end{center}  
\caption{Additional segmentation results on the BSD.}
\label{fig:segresults}
\end{figure}

Next, we present segmentation results on selected images from BSD, 
see Fig.~\ref{fig:segresults}. Images are from both the test and the train set.
Each pair consists of the original image on the left and the segmented image on the right. 
Parameters are $\delta=0.15$, $m=308$, $\epsilon=\frac{\delta}{2}$ $\eta=5$, 
and $n_{max} = 10$. We recall that $m = \frac{8}{15}\pi^2\eta N \delta^5$.
  The selected interaction range results in a relatively small number of 
clusters. Note that, it depends on the type of picture, typically images of nature 
exhibit similar colour, resulting in fewer clusters.

\subsection*{Edge detection}

\begin{figure}[h!]
  \begin{center}
    \includegraphics[width=0.5\textwidth]{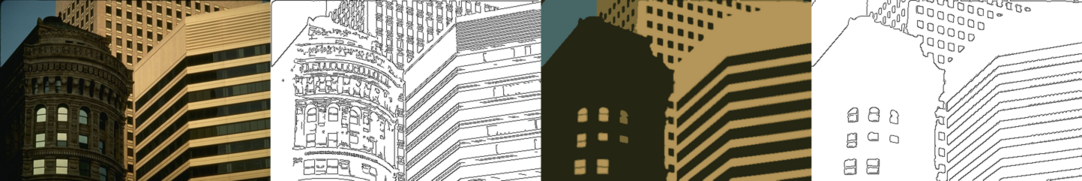}
    \includegraphics[width=0.5\textwidth]{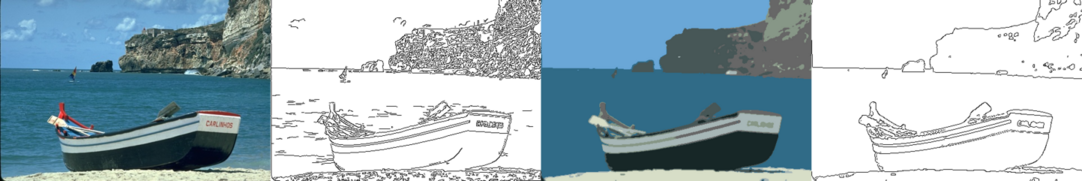}
    \includegraphics[width=0.5\textwidth]{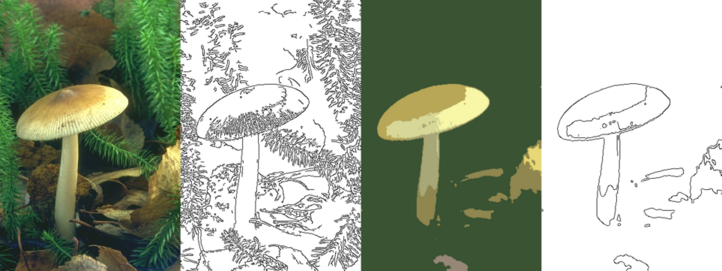}
  \end{center}  
  \caption{Edge detection on original and segmented images.}
  \label{fig:edges}
\end{figure}

Finally, we briefly showcase the impact of our method on segmentation on 
contour detection. In Fig.~\ref{fig:edges} we present, image and edge 
detection for input (left) and segmented (right) image, respectively. 
Contours are detected by the method of Canny \cite{Canny1986}. 
We observe, the improvement of edge detection for the segmented image compared to the direct 
application of Canny method to the original image.

\section{Mathematical proofs}\label{sec:math}

In this section we prove Theorem \ref{theo1}. The starting point is Lemma 
\ref{con}, which connects the notion of $r$-densely packed clusters with their 
connectivity. We recall the notation established in Section  \ref{sec:dynamics}.

Throughout this section we denote $$\vx:=(x_1,...,x_N)\in\R^{dN}.$$

\smallskip
\begin{lem}\label{con}
{\it
Suppose that $\vx$ is a smooth solution to \eqref{1st} with constant 
coefficients in $[t_0,t_1)$. Assume further that $\A$ is an isolated $r$-densely packed 
cluster at each $t\in[t_0,t_1)$ for $r\leq \delta$. Then 
    \begin{enumerate}
        \item a cluster $\A$ treated as a graph is 
        strongly connected and undirected,
        \item The graph diameter $d_\A$ of cluster $\A$, {\rm i.e.} the maximal 
        shortest path between any two nodes, is upper-bounded by
    \begin{equation}
    d_\A\leq
    \begin{cases}
        \# \A & \mbox{if}\ m=1,\\
        \left[\frac{3\#\A}{m+1}\right]-1 & \mbox{if}\ m\geq 2.
    \end{cases}
\end{equation}
\item The spatial diameter ${\mathcal D}_\A:= \max\{|x_i-x_j|:i,j\in\A\}$ 
of~a~cluster $\A$ is upper-bounded by
\begin{align*}
    {\mathcal D}_\A\leq rd_\A.
\end{align*}
    \item There exists $\lambda>0$ such that
    \begin{align*}
    |x_i(t)\!-\!x_\A|\leq e^{-\lambda (t-t_0)}|x_i(t_0)\!-\!x_\A|,\, \forall i\in 
    \A ,
     \, t\in[t_0,t_1),
    \end{align*}
    where $x_\A$ is the center of mass of $\A$. Moreover the~exponent $\lambda$ 
    satisfies
    \begin{align}\label{beq}
        \lambda\geq \frac{4\kappa}{d_\A\#\A}.
    \end{align}
    \end{enumerate}
}
\end{lem}
\smallskip

\begin{proof}
We refer to \cite[Lemma 1 and Lemma 2]{MiMuPe2020}, where variants of assertions 1 and 4 are proved by modifying the theory of linear consensus developed in \cite{OLF1} and \cite{OLF2}. The only 
difference here is that cluster $\A$ is undirected due to the symmetry of 
weight in \eqref{1st}, which further implies that the steady state $x_\A$ is 
indeed the center of mass of $\A$.
Assertion 1 implies that $\A$ is an isolated, connected undirected graph with a 
minimal degree of $m$ and assertion 2 holds true due to Erd\" os, Pach, Pollack 
and Tuza's classical work  \cite[Theorem 1]{ERD}. To prove assertion 3, consider the graph ${\mathcal 
  G}:=(\A,{\mathcal E}_r)$, with ${\mathcal E}_r(i,j)=1$ iff ${\mathcal 
  E}(i,j)=1$ and $|x_i-x_j|\leq r$. Then ${\mathcal G}$ is an $r$-densely 
packed subgraph of $(\A,{\mathcal E})$ with the same nodes and fewer edges. Any two nodes $i$ and $j$ in $\A$ 
are connected by a path $\pi(i,j)\subset {\mathcal E}_r$ represented by a sequence of nodes $i= 
k_1,...,k_l = j$ with
\begin{align*}
    l\leq d_\A\,\,\mbox{and}\,\, |x_{k_\alpha}-x_{k_{\alpha+1}}|\leq r\ 
    \,\,\mbox{for all}\,\, \alpha\in\{1,...,l-1\}.
\end{align*}
Consequently, $|x_i-x_j|\leq rd_\A$.

Finally, since $\lambda(t)$ is the algebraic connectivity of the 
graph $(\A,{\mathcal E})$ multiplied by $\kappa$, inequality \eqref{beq} 
follows from McKay's result published by Mohar in \cite[Theorem 4.2]{MOHAR}, which states that the 
algebraic connectivity of the graph is lower-bounded by $4$ divided by the 
graphs diameter and its number of nodes.
\end{proof}
We proceed with the following useful proposition.

\smallskip
\begin{prop}\label{prop1}
{\it
For any smooth solution $\vx$ to system \eqref{1st} and all $t\geq t_0\geq 0$ we have
\begin{align}\label{maxex}
    {\mathcal C}(t):={\rm conv}\{x_i(t): i\in\{1,...,N\}\}\subset {\mathcal 
      C}(t_0).
\end{align}
Moreover the maximal velocity of each agent is uniformly bounded, {\rm i.e.}
\begin{align}\label{maxv}
   V_i(t):=\max_{t\geq t_0}|\dot{x}_i(t)|\leq \kappa\#\cN_i\delta. 
\end{align}
}
\end{prop}
\smallskip

\begin{proof}  
 Fix any $s\geq t_0$ and let $x_i(s)$ belong to the boundary of ${\mathcal C}(s)$ and assume without a loss of generality that $x_i(s)=0$. It suffices to show that $\dot{x}_i(s)$ belongs to the cone $\bigcup_{\tau>0}\tau{\mathcal C}(s)$. With $\chi_{\{k\in\cN_i\}}$ denoting the characteristic function of the event that $k\in\cN_i(t)$, we have
 \begin{equation}\label{conv}
  \begin{split}
     \dot{x}_i(s) &= \kappa\sum_{k\in\cN_i}(x_k(s)-x_i(s)) \\ &= 
     \kappa\#\cN_i\sum_{k=1}^N\frac{\chi_{\{k\in\cN_i\}}}{\#\cN_i}x_k(s)\in 
     \kappa\#\cN_i{\mathcal C}(s),
 \end{split}
\end{equation}
 since the sum on the right-hand side of \eqref{conv} is a convex 
 combination of elements belonging to ${\mathcal C}(s)$.
 
The proof of \eqref{maxv} follows immediately from  \eqref{1st} and \eqref{nor} after noting that $|x_i-x_k|\leq \delta$ for $k\in \cN_i$.
\end{proof}

The following lemma is the essential part of the proof of Theorem~\ref{theo1}.

\smallskip
\begin{lem}\label{main21}
{\it
Let $\vx$ be a smooth solution to \eqref{1st} and let $\A$ be a cluster in $\{1,...,N\}$ that remains isolated at all times. Then there exists $r^*<\delta$ 
such that if ${\mathcal A}$ is $r$-densely packed with $r\leq r^*$, then 
${\mathcal A}$ is at least $\delta$-densely packed for all $t\geq t_0$.
}
\end{lem}
\smallskip

\begin{proof}
Any $r$-densely packed cluster is $r^*$-densely packed whenever $r\leq r^*$, hence for the remainder of the proof we will assume that $r=r^*$. Since, at $t=t_0$, the agents are $r$-densely packed for $r<\delta$ and the 
maximal velocity of the agents is uniformly bounded (see Proposition 
\ref{prop1}), there exists a time interval $[t_0,T)$, such that the 
agents are $\delta$-densely packed for $t\geq t_0$.

Cluster $\A$ is isolated and system \eqref{1st} is piecewise linear with finitely many possible right-hand 
 sides, and thus,  Lemma~\ref{con} holds locally at each $t\geq t_0$.  Therefore 
 there exists a piecewise constant exponent $\lambda(t)>0$ as in assertion 4 of Lemma~\ref{con}. By assertion 2 as well as inequality~\eqref{beq} from Lemma~\ref{con} the exponent has the uniform lower bound 
\begin{align*}
    \lambda(t)\geq \frac{4\kappa}{d_\A\#\A}\geq \frac{4\kappa m}{3(\#\A)^2} =: \lambda_*.
\end{align*}
 Therefore for all $t\geq t_0$ and all $i\in\A$ we have
\begin{equation}\label{exp}
\begin{split}
    |x_i(t)-x_\A(t)|\leq e^{-\lambda_* (t-t_0)}{\mathcal D_\A}(t_0),\\ 
    {\mathcal D_\A}(s) := \sup_{i,j\in\A}|x_i(s)-x_j(s)|.
\end{split}
\end{equation}

 Let
    \begin{align*}
    T := \sup\{t>t_0: \A \mbox{ is } \delta\mbox{-densely packed in }[t_0,t)\}>0.
    \end{align*}
    
    We shall chose $r$ small enough so that $T=\infty$. For all $i,j\in{\mathcal A}$ and all $t\in[t_0,T)$, we have

    \begin{equation}\label{cin}
    |x_i(t)-x_j(t)| \leq |x_i(t)-x_\A(t)| + |x_j(t)-x_\A(t)|.
    \end{equation}

Let us fix in \eqref{cin} any pair $(i,j)$ such that $|x_i(t_0)-x_j(t_0)|\leq r$. Such pairs exist since the ensemble is $r$-densely packed initially. Applying \eqref{exp} to the right-hand side of \eqref{cin} leads to
\begin{align}\label{pom1}
|x_i(t)-x_j(t)|\leq 2e^{-\lambda_* (t-t_0)}{\mathcal D_\A}(t_0).
\end{align}
Alternatively $|x_i(t)-x_j(t)|$ can be upper-bounded using equation \eqref{1st} and Proposition \ref{prop1} yielding
\begin{align*}
    |x_i(t)-x_j(t)|\leq|x_i(t)-x_i(t_0)| + |x_j(t)-x_j(t_0)| \\+ 
    |x_i(t_0)-x_j(t_0)|\leq 2\kappa\#\A\delta (t-t_0)  + r,
\end{align*}
which ensures that
\begin{align*}
    |x_i(t)-x_j(t)|< \delta
\end{align*}
at least as long as
\begin{align*}
t-t_0< \frac{1}{2\kappa\#\A}\left(1-\frac{r}{\delta}\right).
\end{align*}
However for $t-t_0\geq \frac{1}{2\kappa\#\A}\left(1-\frac{r}{\delta}\right)$, by \eqref{pom1} we have
\begin{align*}
    |x_i(t)-x_j(t)|\leq 2e^{-\frac{\lambda_*}{2\kappa\#\A}\left(1-\frac{r}{\delta}\right)}{\mathcal D_\A}(t_0)
    \end{align*}
    and we require
    \begin{align}\label{d1}
\tilde{r}:=        2e^{-\frac{\lambda_*}{2\kappa\#\A}\left(1-\frac{r}{\delta}\right)}{\mathcal D_\A}(t_0)< \delta.
    \end{align}
Then assertions 2 and 3 of Lemma \ref{con} imply that
\begin{align*}
    {\mathcal D}_\A< 3r\frac{\#\A}{m}.
\end{align*}
Combining the above bounds yields
\begin{align*}
\tilde{r}<  6re^{-\frac{2 m}{3(\#\A)^3}(1-\frac{r}{\delta})}\frac{\#\A}{m}\leq \delta,
\end{align*}
which can be further rearranged  into \eqref{warunek}. Condition \eqref{warunek} holds for any sufficiently small $r>0$.

In conclusion, any pair $i$ and $j$, initially of distance at most $r$ from each other, remains of distance $\tilde{r}<\delta$ from each other throughout $[t_0,T)$. Thus $\A$ is $\tilde{r}$-densely packed in  $[t_0,T)$ with $\tilde{r}<\delta$ and, by continuity, it is at least $\frac{\tilde{r}+\delta}{2}$-densely packed at $t=T$. Then, exactly as at the beginning of the proof, $\A$ remains at least $\delta$-densely packed past the time $T$ which stands in contradiction with the definition of $T$, unless $T=\infty$. The proof is finished.
\end{proof}

Lemma \ref{main21} ensures that $\A$ remains $\delta$-densely packed as long as \eqref{warunek} holds and $\A$ is an isolated cluster. We are now ready to finalise the proof of Theorem \ref{theo1}.

{\it Proof of Theorem \ref{theo1}:}
First observe that, by Proposition \ref{prop1}, condition \eqref{wypuk} persists in time. Hence, each of the clusters $\A_1,...,\A_K$ is isolated, and thus Lemma \ref{main21} applies with $r^*$ satisfying condition \eqref{warunek}. Therefore, each $r^*$-densely packed cluster is at least $\delta$-densely packed indefinitely. Consequently, for such clusters, assumptions of Lemma \ref{con} are satisfied, and inequality \eqref{exp} holds for all $t\geq t_0$ thereby ensuring the convergence in Theorem \ref{theo1}. The proof is finished. $\hfill\rule{1.4ex}{1.4ex}$

\section{Conclusions and Outlook}\label{sec:conc}

The paper explores the relation between density-based data segmentation and 
collective behavior, augmenting the DBSCAN, by letting the data evolve in 
accordance with a non-local interaction law. 
We provided a rigorous mathematical foundation for the method as well as an illustration
in the case of color image segmentation. 
Various aspects of the algorithm are discussed: the influence of its 
parameters, 
stopping time and numerical complexity. In particular, we achieve a low $O(N)$ 
average numerical complexity with parameters that lead to the emergence of an 
appropriate number of clusters. The key point is the observation that an evolution of the data according to
a density-based first-order ODE system has a low computational cost, and breaks unreasonably large 
clusters that would be obtained using other methods such as DBSCAN.

Future research will be dedicated to the connections between collective 
dynamics and more modern clustering techniques.

\section*{Acknowledgment}

PM acknowledges the financial support by the Federal Ministry of 
 Education and Research of Germany, grant number 05M16NMA and the support of 
the GRK 2297 MathCoRe, funded by the Deutsche Forschungsgemeinschaft, grant 
 number 314838170. JP was partially supported by the Polish Narodowe Centrum Nauki grant No. 2018/30/M/ST1/00340  (HARMONIA).


\end{document}